\documentclass[journal,twoside,web]{ieeecolor}
\usepackage{generic}
\usepackage{xcolor}
\usepackage{cite}
\usepackage{amsmath, amssymb, algorithm, algpseudocode}
\usepackage{graphicx}
\usepackage{algorithm}
\usepackage{hyperref}
\hypersetup{hidelinks=true}
\usepackage{textcomp}

\usepackage{multirow}
\usepackage{booktabs}

\usepackage{subfigure}
\usepackage{tabularx}
\usepackage{threeparttable} 
\usepackage{makecell}

\usepackage{siunitx}
\def\BibTeX{{\rm B\kern-.05em{\sc i\kern-.025em b}\kern-.08em
    T\kern-.1667em\lower.7ex\hbox{E}\kern-.125emX}}
\begin{document}
\title{SASG-DA: Sparse-Aware Semantic-Guided Diffusion Augmentation For Myoelectric Gesture Recognition}
\author{Chen Liu, Can Han, Weishi Xu, Yaqi Wang, and Dahong Qian, \textit{Senior Member, IEEE}
\thanks{© 2026 IEEE. The final published version is available at https://doi.org/10.1109/JBHI.2026.3670388.}
\thanks{This work was supported in part by the Fundamental Research Funds for the Central Universities (YG2024LC10).}
\thanks{Chen Liu, Can Han, Weishi Xu, and Dahong Qian (Corresponding author) are with the School of Biomedical Engineering, Shanghai Jiao Tong University, Shanghai 200240, China. (e-mail: lchen1206@sjtu.edu.cn.)}
\thanks{Yaqi Wang is with the Innovation Center for Electronic Design Automation Technology, Hangzhou Dianzi University, Hangzhou, 310018, China.}
}

\maketitle

\begin{abstract}
Surface electromyography (sEMG)-based gesture recognition plays a critical role in human–machine interaction (HMI), particularly for rehabilitation and prosthetic control. However, sEMG-based systems often suffer from the scarcity of informative training data, leading to overfitting and poor generalization in deep learning models. Data augmentation offers a promising approach to increasing the size and diversity of training data, where faithfulness and diversity are two critical factors to effectiveness. However, promoting untargeted diversity can result in redundant samples with limited utility. To address these challenges, we propose a novel diffusion-based data augmentation approach, Sparse-Aware Semantic-Guided Diffusion Augmentation (SASG-DA). To enhance generation faithfulness, we introduce the Semantic Representation Guidance (SRG) mechanism by leveraging fine-grained, task-aware semantic representations as generation conditions. To enable flexible and diverse sample generation, we propose a Gaussian Modeling Semantic Sampling (GMSS) strategy, which models the semantic representation distribution and allows stochastic sampling to produce both faithful and diverse samples. To enhance targeted diversity, we further introduce a Sparse-Aware Semantic Sampling strategy to explicitly explore underrepresented regions, improving distribution coverage and sample utility. Extensive experiments on benchmark sEMG datasets, Ninapro DB2, DB4, and DB7, demonstrate that SASG-DA significantly outperforms existing augmentation methods. Overall, our proposed data augmentation approach effectively mitigates overfitting and improves recognition performance and generalization by offering both faithful and diverse samples.
\end{abstract}


\section{Introduction}
\label{sec:introduction}
\IEEEPARstart{G}esture recognition serves as a fundamental technology for advancing human–machine interaction (HMI). Among various gesture recognition modalities, surface electromyography (sEMG)-based approaches have gained increasing attention due to their non-invasive nature, high temporal resolution, and ability to directly capture muscle activation signals associated with voluntary movement~\cite{kaifosh2025generic}. sEMG sensors can be conveniently worn on the skin surface, making them well-suited for continuous and user-friendly deployment in real-world scenarios~\cite{sivakumar2024emg2qwerty}, particularly in fields such as rehabilitation training, prosthetic control, and wearable human–machine interfaces.

\begin{figure}[!t]
    \centering
    \includegraphics[width=0.95\columnwidth]{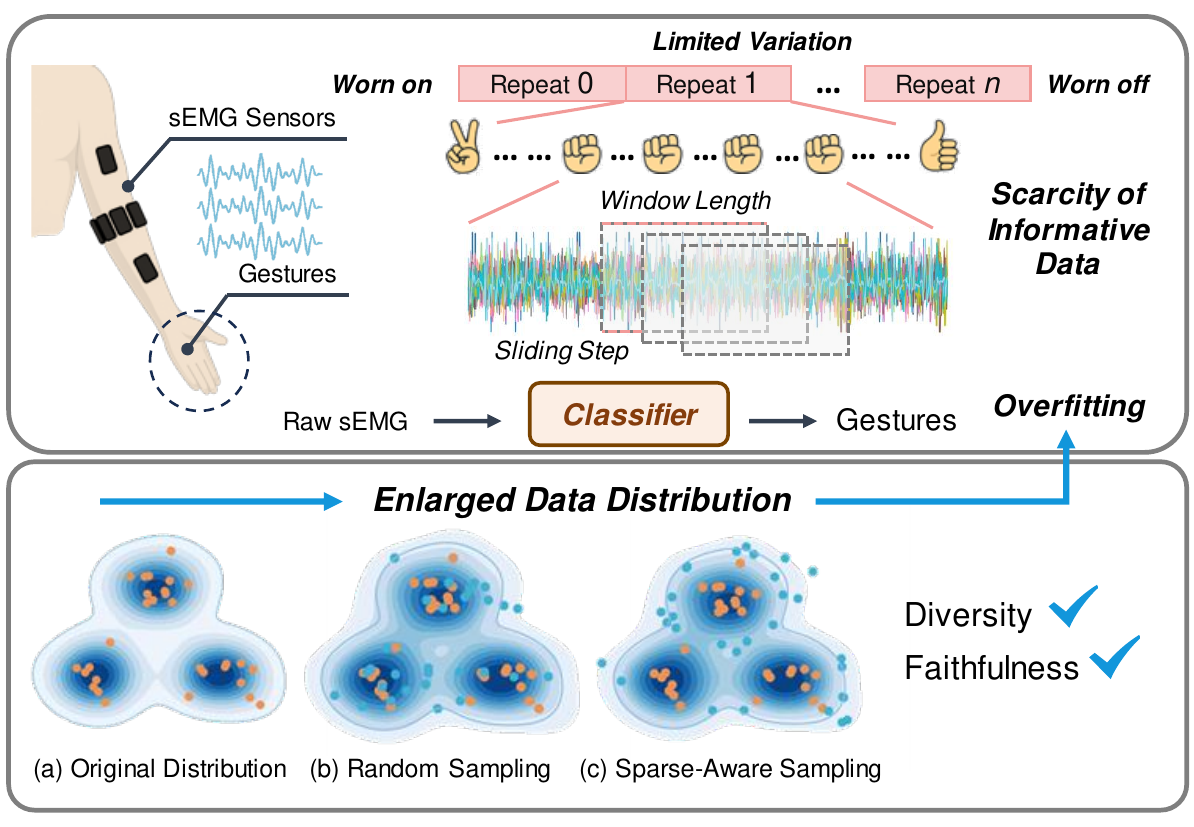}
     \vspace{-0.3cm}
    \caption{Overfitting remains a persistent challenge in sEMG-based gesture recognition due to the scarcity of informative data. We alleviate this by generating faithful and diverse samples through diffusion-based augmentation. Our sparse-aware sampling strategy further encourages generation in underrepresented regions, effectively expanding the data distribution.}
    \vspace{-0.7cm}
    \label{motivation}
\end{figure}

While deep learning has significantly improved gesture recognition performance, sEMG-based systems remain highly prone to overfitting due to the scarcity of informative training data. This scarcity arises from two main factors: the inherent difficulty in collecting large-scale sEMG datasets, and the prevalence of redundant information within the recorded data (Fig.~\ref{motivation}). On the one hand, sEMG acquisition is labor-intensive and time-consuming, requiring substantial human and annotation costs, which constrain the overall dataset size. On the other hand, common collection protocols often involve subjects repeating each gesture multiple times within a single session. However, due to the limited variation across repetitions, the resulting data tend to be homogeneous. Moreover, sliding-window segmentation during preprocessing further introduces redundancy, as overlapping windows generate numerous similar samples. Therefore, increasing the size and diversity of available data is essential for improving the performance and generalization of sEMG-based gesture recognition systems.

Data augmentation offers a promising approach to increasing the size and diversity of training data. Previous studies on sEMG-based gesture recognition have explored both single-sample augmentation~\cite{tsinganos2020data} and GAN-based augmentation~\cite{ao2024overcoming}, which often suffer from limited diversity and unstable generation, respectively. More recently, diffusion-based augmentation has attracted significant interest for its ability to generate faithful and diverse samples~\cite{dhariwal2021diffusion}. Nevertheless, despite its success in computer vision, only PatchEMG~\cite{xiong2024patchemg} has explored this direction in sEMG-based gesture recognition, focusing on generating high-quality data in a few-shot setting. In contrast, our work highlights two critical aspects of data augmentation: faithfulness and diversity~\cite{wang2024enhance, wang2025inversion, trabucco2024effective}. Faithfulness ensures that generated samples remain semantically consistent with the distribution of the original data, enabling the downstream model to learn from correct augmented data~\cite{wang2025inversion}, while diversity enriches the original data distribution and helps the model learn from robust and invariant features from diverse samples~\cite{wang2024enhance, wang2025inversion}. 
For instance, DiffMix~\cite{wang2024enhance} emphasizes introducing a diverse background while maintaining a faithful foreground, whereas DA-Fusion~\cite{trabucco2024effective} considers diversity in the textual prompt space. However, direct application of these approaches to sEMG signals remains limited due to the lack of explicit textual prompts and a clear foreground-background structure.
Furthermore, promoting untargeted diversity can result in redundant samples that offer limited incremental value, an aspect that only a few studies~\cite{li2025gendataagent} have explored. Specifically, when diversity is introduced without considering the semantic coverage of the training set, the generated samples may cluster around already well-represented regions in the feature space, potentially reducing sample utility. Therefore, ensuring both faithfulness and targeted diversity helps expand the data distribution with informative samples, thereby improving the performance of sEMG-based gesture recognition systems (Fig.~\ref{motivation}).

To this end, we propose a novel diffusion-based augmentation method, Sparse-Aware Semantic-Guided Diffusion Augmentation (SASG-DA) for sEMG-based gesture recognition, which aims to generate both faithful and diverse augmented samples. To enhance generation faithfulness, we introduce a Semantic Representation Guidance (SRG) mechanism, which leverages fine-grained, class-specific representations as semantic conditions to guide the diffusion training process. These semantic representations capture rich contextual information, enabling the model to generate synthetic samples that are not only realistic but also well-aligned with the target classes. Building upon SRG, we further propose a Gaussian Modeling Semantic Sampling (GMSS) mechanism for the diffusion inference process to enable flexible and diverse sample generation. GMSS introduces diversity by stochastically sampling novel semantic conditions for generation. However, untargeted diversity may lead to redundant samples. To this end, we further extend the GMSS by introducing a Sparse-Aware Semantic Sampling (SASS) mechanism that actively targets sparse regions in the semantic space for conditional sampling, which explicitly expands the coverage of the training distribution. In this setting, diversity arises not only from the inherent stochasticity of the diffusion process but also from the stochastic variation introduced in the semantic condition space. Through the integrated strategy of SRG and SASS, SASG-DA offers a principled augmentation method that enhances faithfulness and diversity, effectively mitigating overfitting and improving the performance and generalization of sEMG-based gesture recognition.

{\bf Summary of Contributions}
\begin{itemize}
    \item [1)] We propose a novel diffusion‑based data augmentation method, Sparse-Aware Semantic-Guided Diffusion Augmentation (SASG-DA), for sEMG‑based gesture recognition. The proposed method can provide both faithful and diverse augmented samples to mitigate the scarcity of informative training data.  
    \item [2)] To enhance generation faithfulness, we propose a Semantic Representation Guidance (SRG) mechanism which leverages fine‑grained semantic representations as conditions, enabling the downstream model to learn from accurate and class-consistent information.
    \item [3)] To enhance targeted generation diversity, we propose a Sparse‑Aware Semantic Sampling (SASS) mechanism that actively focuses on underrepresented regions in the semantic space, explicitly expanding the training distribution and producing more informative, diverse augmented samples.
    \item [4)] We comprehensively validate SASG-DA on three benchmark sEMG datasets, Ninapro DB2, DB4, and DB7, showing significant improvements over state-of-the-art methods and demonstrating the effectiveness of our augmentation approach in enhancing the performance and generalization.

\end{itemize}

\section{Related works}
\subsection{sEMG-based Gesture Recognition}
Surface electromyography (sEMG) signals reflect muscle activation patterns and have been widely used for gesture recognition, especially in human–computer interaction and rehabilitation applications~\cite{kaifosh2025generic}. Recent advances in deep learning have significantly improved sEMG-based gesture recognition by enabling end-to-end learning from raw signals. Numerous studies have focused on designing tailored network architectures, such as Spatio-Temporal Cross Networks (STCNet)~\cite{yang2025stcnet}, Kolmogorov-Arnold Network~\cite{al2024tcnn}, and Transformer-based models~\cite{wang2024transformer}, to enhance recognition accuracy. These models aim to capture the complex, non-stationary patterns of sEMG signals to enhance recognition performance and generalization. In contrast, our work focuses on data augmentation as an orthogonal approach to enhance model performance.

\subsection{Data Augmentation for sEMG-based Gesture Recognition}

Data augmentation is crucial in improving generalization and mitigating overfitting, particularly with limited training data. Existing sEMG augmentation strategies can be broadly categorized into single-sample DA and generative-based DA.

Single-sample sEMG augmentation methods~\cite{tsinganos2020data} apply various transformations to individual data instances while preserving labels, typically in the time or frequency domain~\cite{zhao2024dominant}. 
Although single-sample DA methods are simple and computationally efficient, they often offer limited diversity and insufficient variations for enhancing model generalization~\cite{cao2024survey}.

Generative models provide a more principled approach to data augmentation by learning the underlying data distribution and synthesizing entirely new samples. Previous sEMG augmentation studies based on GANs~\cite{ao2024overcoming} have demonstrated their effectiveness in mitigating data scarcity and improving robustness. Building upon these advancements, diffusion-based models have recently emerged as a compelling alternative, offering superior generation ability in terms of both sample diversity and faithfulness~\cite{dhariwal2021diffusion}. However, only a few studies have explored their application to sEMG-based gesture recognition. PatchEMG~\cite{xiong2024patchemg} is the first to apply diffusion models to few-shot sEMG signal synthesis, focusing on generating high-quality data from limited samples and thereby improving gesture recognition accuracy. Specifically, it applies a patch-based training strategy to address the challenge of limited data for diffusion training, and during inference it generates complete-length signals and employs classifier-free guidance (CFG)~\cite{ho2021classifier} to ensure generation quality.

Given that EEG and EMG are both low-SNR, stochastic, and non-stationary biosignals~\cite{luo2025diffusion}, several representative EEG-oriented diffusion methods are conceptually relevant to sEMG data augmentation, including SSE-LDM~\cite{aristimunha2023synthetic}, EEG-Enhancing~\cite{siddhad2024enhancing}, NTD~\cite{vetter2024generating} and DESAM~\cite{luo2025diffusion}. Among them, DESAM~\cite{luo2025diffusion} focuses on enhancing the faithfulness of generated samples through temporal mixup with real data. While these studies focus on improving the modeling and generation quality of diffusion models for EEG signals, they have not fully explored how to generate more effective augmented samples, relying mainly on the inherent diversity of diffusion models.

To the best of our knowledge, PatchEMG~\cite{xiong2024patchemg} and our work are among the first attempts that leverage diffusion models for sEMG data augmentation. While PatchEMG focuses on the few-shot setting, our work explores a broader augmentation approach to improve the downstream effectiveness of sEMG-based gesture recognition by emphasizing two critical factors: generation faithfulness and diversity.

\subsection{Diffusion Model Generation}
Due to the various requirements of downstream tasks, diffusion models are designed with different generation objectives. Data augmentation, diverse generation, and minority sample generation are the most relevant to our work.

Recent diffusion-based augmentation methods primarily aim to improve the faithfulness and diversity of generated samples~\cite{wang2024enhance, wang2025inversion, trabucco2024effective}. For instance, DiffuseMix~\cite{islam2024diffusemix} preserves key semantics by concatenating generated and original samples using a binary mask. Other methods emphasize enhancing diversity of generated samples~\cite{wang2024enhance, trabucco2024effective, rahat2025data}. Among them, DiffMix~\cite{wang2024enhance} interpolates inter-class images during inference to increase diversity, producing novel counterfactual samples such as land birds in maritime settings; DA-Fusion~\cite{trabucco2024effective} boosts diversity by inserting and fine-tuning new tokens in the text encoder to capture novel visual concepts. 
Beyond these efforts, only a few studies have explored the practical utility of augmented samples in downstream tasks~\cite{li2025gendataagent}. As untargeted diversity may lead to redundant or typical samples, GenDataAgent~\cite{li2025gendataagent} incorporates downstream feedback to prioritize the generation of challenging samples. However, such feedback-driven approaches are often constrained by the slow inference speed of diffusion models. Moreover, most diffusion-based image augmentation methods are not readily applicable to sEMG signals due to their modality-specific design.  

A related but distinct objective is enhancing the diversity of diffusion generation, which has received comparatively less attention than improving generation quality~\cite{sadat2024cads}. Notable progress was made in CADS~\cite{sadat2024cads}, which demonstrated that gradually annealing noise perturbations to conditions can significantly boost the diversity of generated samples. However, this approach is not intended for data augmentation.

Minority sample generation targets low-density regions of the data distribution, since diffusion model samplers inherently favor high-density regions of the data manifold~\cite{sehwag2022generating}. Existing approaches typically address this limitation through gradient-based guidance, leveraging metrics such as the Hardness Score~\cite{sehwag2022generating} or Minority Score~\cite{um2024self} to generate minority samples. However, such gradient-guided methods often rely on additional external components and incur substantial computational overhead~\cite{um2024self}.

Distinct from prior works that focus solely on diverse or minority sample generation, our approach simultaneously emphasizes the faithfulness and diversity of generated data, which is crucial for effective data augmentation. Furthermore, our approach incorporates the Sparse-Aware Semantic Sampling (SASS) mechanism that explicitly expands the original data distribution to enhance augmentation utility, rather than focusing solely on rare samples in the target scenario. The latter approaches may generate data that deviates from the true distribution, which may limit their utility for augmentation. Moreover, our SASS mechanism introduces minor cost while requiring no additional components or gradient guidance.

\begin{figure*}[!ht]
    \centering
        \includegraphics[width=0.83\textwidth]{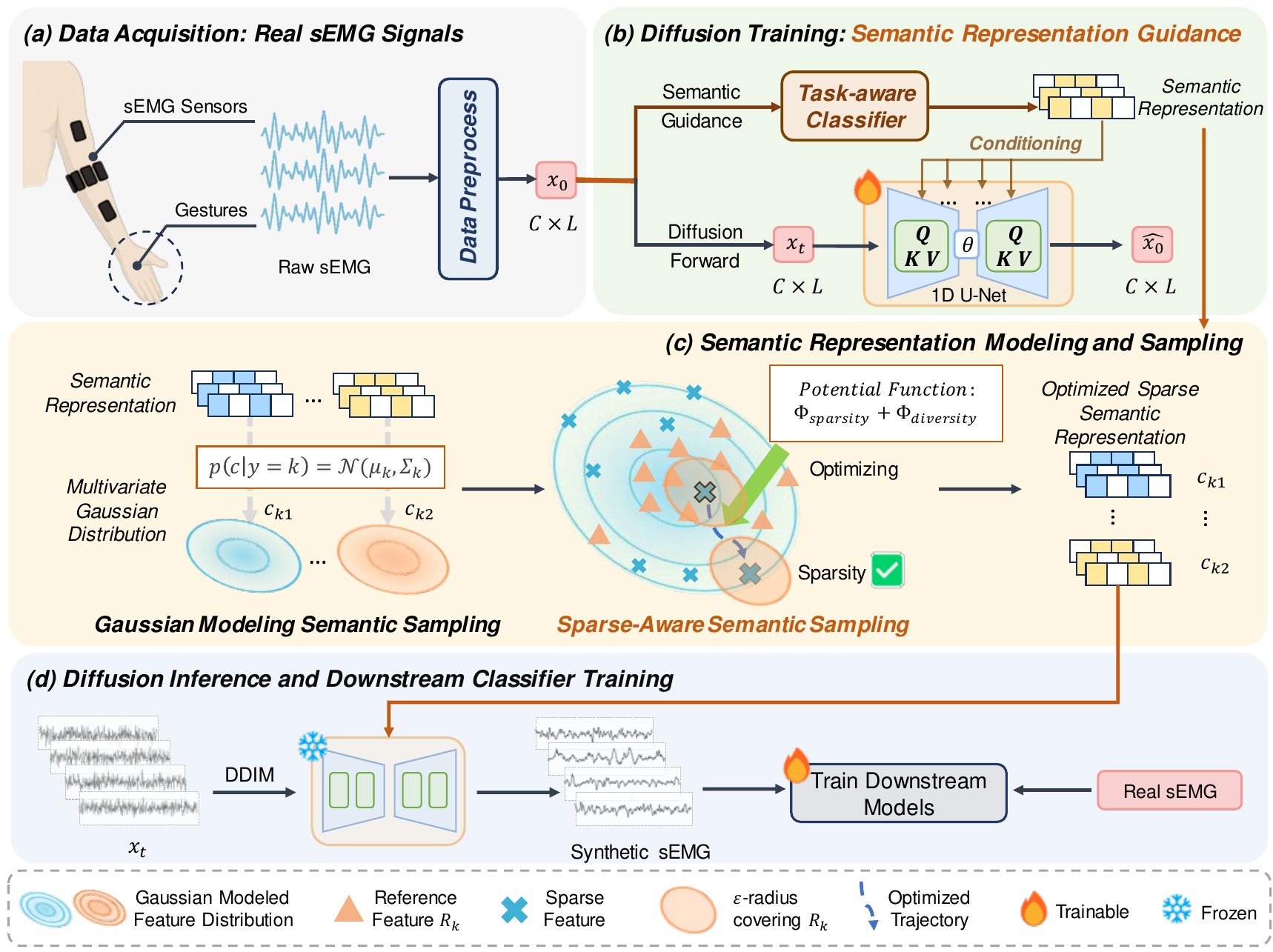}
    \vspace{-0.2cm}
    \caption{Overview of the proposed SASG-DA approach based on the diffusion model. (b) During diffusion training, SASG-DA enhances generation faithfulness by introducing semantic representation guidance. (c) During inference, SASG-DA models the semantic representation distribution with a Gaussian and performs Sparse-Aware Semantic Sampling to explicitly target sparse regions in the semantic space, thereby expanding the data distribution. (d) The optimized sparse semantic representations are then used as conditions for diffusion reverse and generate synthetic sEMG signals for downstream model training.}
    \label{fig: framework}
    \vspace{-0.6cm}
\end{figure*}

\section{Preliminary}
\subsection{Diffusion Models for Data Generation}
Denoising Diffusion Probabilistic Models (DDPMs) approximate complex data distributions by modeling a Markovian forward diffusion process and learning its corresponding reverse denoising process~\cite{ho2020denoising}. In this work, we adapt the DDPM for sEMG generation, aiming to mitigate the scarcity of informative training data. 

The forward process transforms the original data distribution into a standard Gaussian by gradually adding noise. Formally, given an sEMG sample $x_0 \in \mathbb{R}^{C \times L}$, which represents a signal with $C$ channels and $L$ time steps, the noisy sequence $\{x_t\}_{t=1}^T$ is generated by $q(x_t | x_{t-1}) = \mathcal{N}(x_t; \sqrt{1 - \beta_t} x_{t-1}, \beta_t \mathbf{I})$, where $\beta_t$ denotes the noise schedule at timestep $t$. 

The reverse process is a learned Markov chain that progressively denoises a sample from pure Gaussian noise to recover data resembling the original distribution. It is defined by $p_\theta(x_{t-1} | x_t)=  \mathcal{N}(x_{t-1}; \mu_\theta(x_t, t), \Sigma_\theta(x_t, t))$, where $\mu_\theta$ and $\Sigma_\theta$ are parameterized by a neural network $\theta$~\cite{ho2020denoising}.

In practice, it is common to express $\mu_\theta(x_t, t)$ in terms of either the predicted clean data $\hat{x}_0(x_t, t; \theta)$ or the predicted noise $\hat{\epsilon}(x_t, t; \theta)$. Following the Diffusion-TS approach~\cite{yuan2024diffusion}, our model is trained to predict an estimate $\hat{x}_0$ directly. This leads to a simplified training objective known as denoising score matching, formulated as:
\begin{equation}
\mathcal{L}_{\textit{simple}} = \mathbb{E}_{x_0, x, t} \left[ || x_0 - \hat{x}_0(x_t, t; \theta) ||^2 \right],
\label{eq3}
\end{equation}
where $x_t = \sqrt{\bar{\alpha}_t}x_0 + \sqrt{1 - \bar{\alpha}_t}\epsilon$, $ \epsilon \sim \mathcal{N}(0, \mathbf{I})$, and $\bar{\alpha}_t$ is related to the noise schedule at timestep $t$.

To further improve signal generative fidelity, we incorporate the same Fourier-based loss employed in Diffusion-TS~\cite{yuan2024diffusion} to better preserve the spectral structure of sEMG signals:
\begin{equation}
\mathcal{L}_{\textit{Fourier}} = \mathbb{E}_{x_0, x, t} \left[ || \mathcal{FFT}(x_0) -  \mathcal{FFT}(\hat{x}_0(x_t, t; \theta)) ||^2 \right].
\label{eq4}
\end{equation}

To generate data that aligns with specific class semantics, class-conditional diffusion models incorporate label information throughout the denoising process. In our case, the simplified training objective becomes:
\begin{equation}
\mathcal{L}_{\textit{cond}} = \mathbb{E}_{x_0, x, t, y} \left[ || x_0 - \hat{x}_0(x_t, t, y; \theta) ||^2 \right],
\label{eq5}
\end{equation}
where $y$ is the class label. 

Once trained, the model can generate new samples by drawing $ x_t \sim \mathcal{N}(0, \mathbf{I})$ and iteratively applying the reverse transitions.

\section{Methodology}
\subsection{Overview}
To mitigate the scarcity of informative training data commonly encountered in sEMG-based gesture recognition, we propose a novel augmentation method, SASG-DA, based on diffusion models by offering faithful and diverse samples. To enhance generation faithfulness, SASG-DA incorporates a Semantic Representation Guidance (SRG) mechanism, which conditions the diffusion model on fine-grained semantic representations extracted from the training data, ensuring that generated samples are both faithful and class-consistent. During diffusion inference, SRG is embodied through the proposed Gaussian Modeling Semantic Sampling (GMSS) and Sparse-Aware Semantic Sampling (SASS). To enable flexible and diverse generation, GMSS approximates the semantic representation distribution based on a Gaussian assumption. To further explicitly expand the coverage of the training distribution and enhance data utility, we extend the GMSS by designing a Sparse-Aware Semantic Sampling (SASS) mechanism, which encourages sampling from sparse or underrepresented regions of the modeled semantic space. The overall framework is illustrated in Fig.~\ref{fig: framework}. Given a target class, our method first samples semantic representations via SASS to guide the diffusion model in generating new gesture samples. These samples are then added to the training set to boost the downstream performance of gesture recognition.

\subsection{Semantic Representation Guidance}
Faithfulness is a key factor in data augmentation tasks. Specifically, it requires that synthetic samples preserve the distinctive features of their target classes, thereby ensuring that the downstream classifier learns from accurate and class-consistent information~\cite{wang2025inversion}. However, in conventional class-conditioned diffusion models, label conditions alone often provide only coarse-grained guidance, lacking sufficient semantic richness to generate highly faithful samples.

To overcome this limitation, we propose using fine-grained semantic representations as conditioning inputs. These continuous, class-specific features serve as informative guidance signals, allowing the model to generate more semantically aligned and faithful samples. Specifically, semantic representations $f \in \mathcal{R}^{D_c}$ are extracted from a pretrained task-aware classifier and integrated into the diffusion process via cross-attention mechanisms. The pretrained classifier is trained on the same gesture recognition task and dataset as the diffusion model, ensuring that its extracted features are discriminative and task-relevant, thus providing meaningful semantic guidance for generation. We further apply dropout to the semantic features during training to improve robustness and mitigate overfitting to the conditioning input. In addition, label information is retained and incorporated through additive conditioning. The overall training objective combining~\eqref{eq4} and~\eqref{eq5} is as follows:
\begin{equation}
\begin{split}
\mathcal{L}_{\textit{SRG}} = &\mathbb{E}_{x_0, x, t, y, f} \Big[ 
 \left\| x_0 - \hat{x}_0(x_t, t, y, f; \theta) \right\|^2 \\
& + \left\| \mathcal{FFT}(x_0) - \mathcal{FFT}(\hat{x}_0(x_t, t, y, f; \theta)) \right\|^2 
\Big].
\end{split}
\label{eq8}
\end{equation}

\subsection{Gaussian Modeling Semantic Sampling}

In class-conditioned diffusion frameworks, the faithfulness and diversity of the generated samples depend heavily on the design of the conditioning signal. To enable flexible and diverse generation, we propose a simple yet effective Gaussian Modeling Semantic Sampling (GMSS) mechanism. Specifically, we approximate the semantic representation distribution of each class as a multivariate Gaussian, a widely adopted approach in representation learning, which facilitates diverse sampling while maintaining semantic consistency.

Given a labeled training set, we extract semantic representations for all samples using a pretrained task-aware encoder $E(\cdot)$, such as a backbone CNN or transformer trained on the sEMG-based gesture classification task. For each class $k$, we construct its semantic feature set $\mathcal{F}_k = \{f_i = E(x_i)\ |\ y_i = k\}$ and estimate its empirical mean and covariance:
\begin{equation}
\mu_k = \frac{1}{|\mathcal{F}_k|} \sum_{f \in \mathcal{F}_k} f, \quad 
\Sigma_k = \frac{1}{|\mathcal{F}_k| - 1} \sum_{f \in \mathcal{F}_k} (f - \mu_k)(f - \mu_k)^\top.
\label{eq9}
\end{equation}

We then model the semantic representation distribution as a multivariate Gaussian $\mathcal{N}(\mu_k, \Sigma_k)$. During diffusion inference, a new semantic representation $\tilde{f}_k$ can be sampled as:
\begin{equation}
\tilde{f}_k \sim \mathcal{N}(\mu_k, \Sigma_k).
\end{equation}

The sampled feature $\tilde{f}_k$ is used as a continuous condition to guide the diffusion reverse process, generating semantically aligned yet diverse samples for class $k$.

Compared to discrete and coarse-grained label-based conditioning, GMSS provides a continuous, distribution-based conditioning space that enables flexible sampling within each class. By sampling from different regions of the semantic space, GMSS expands coverage beyond the original training set while preserving semantic faithfulness. This not only increases intra-class diversity but also expands the effective data manifold, thereby improving the generalization performance of classifiers. Thus, diversity arises not only from inherent diffusion stochasticity but also from the stochastic variation introduced in the semantic condition space.

\subsection{Sparse-Aware Semantic Sampling}

While GMSS allows class-consistent sampling in a continuous semantic representation space, the diffusion model inherently tends to generate samples in dense regions of the feature distribution, thereby failing to explore underrepresented or sparsely populated areas~\cite{sehwag2022generating}. These dense regions correspond to well-represented and frequently observed samples in the training data, whereas the sparse regions contain rare or insufficiently represented cases that the model has rarely seen but are crucial for improving downstream performance. To address this limitation, we propose a Sparse-Aware Semantic Sampling (SASS) strategy that explicitly guides the generation process toward sparse regions in the data distribution. SASS employs a global–local sampling strategy: at the global level, we identify candidate features that are likely to lie in sparse regions of the distribution; at the local level, we optimize these candidates using a potential function that jointly encourages sparsity and diversity.

Specifically, we first generate an oversampled set of features $\tilde{f}$ from the class-specific Gaussian distribution $\mathcal{N}(\mu_k, \Sigma_k)$, denoted as the candidate set $C_{k}$. This random oversampling retains class consistency while providing a broad coverage of the semantic representation space.

To identify potential sparse candidates from the oversampled pools, we adopt the \textit{rarity score} proposed in~\cite{han2024rarity}. For each class, the original training samples $R_k$ serve as the reference set to construct $k$-NN spheres, where both the centers and their neighbors are reference samples. The \textit{rarity score} of a candidate is defined as the radius of the smallest sphere that contains it. Intuitively, candidates in dense regions fall within smaller spheres, while those near sparse regions yield higher \textit{rarity scores}. We then rank candidates by their \textit{rarity scores} and select high-rarity ones as the preliminary sparse condition set $\tilde{C}_k$.

\begin{figure}[!t]
    \centering
        \includegraphics[width=\columnwidth]{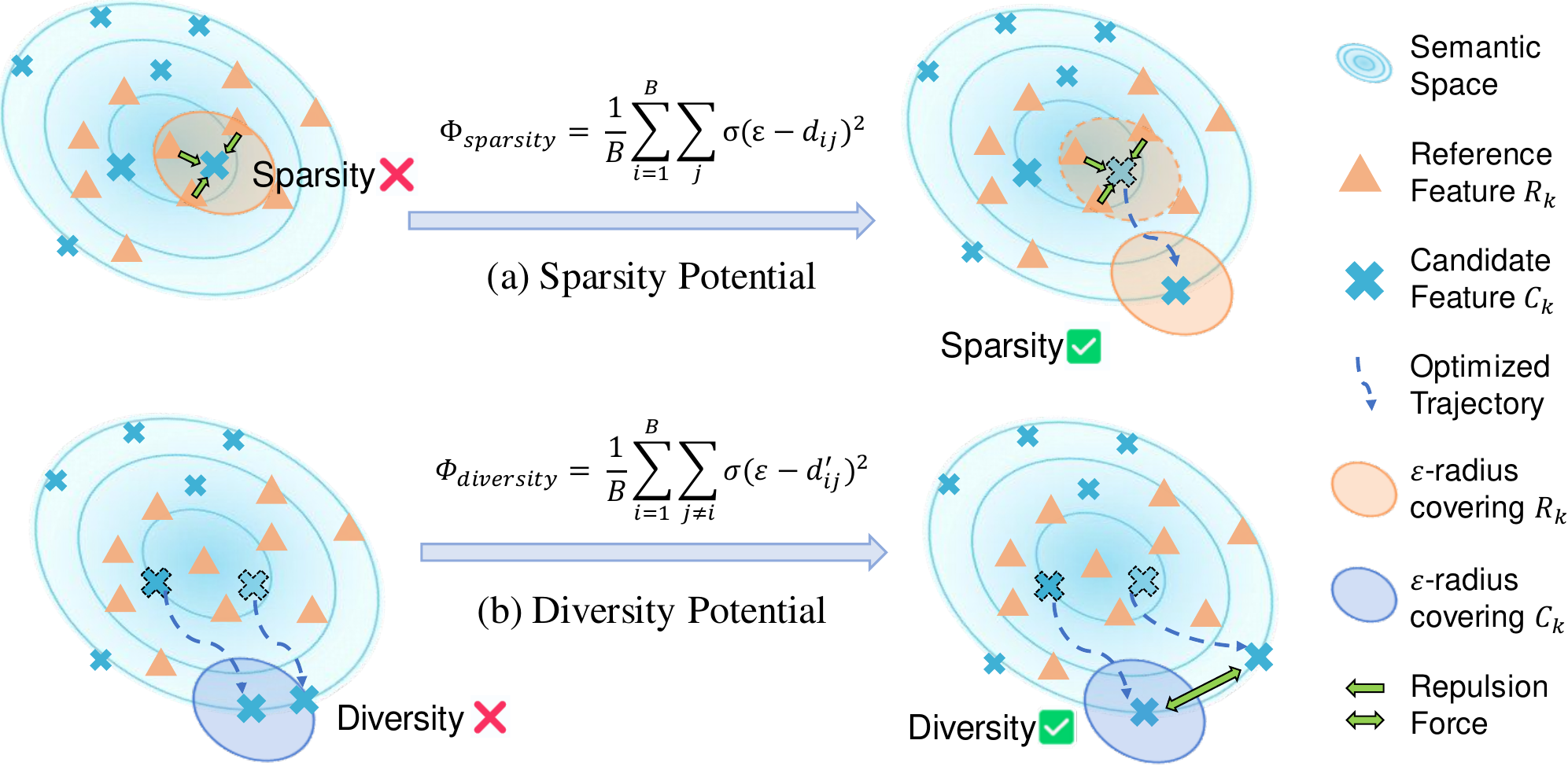}
    \caption{Sparse-Aware Semantic Sampling. Candidate sparse features (blue) are optimized in the semantic space using two potentials: (a) a sparsity potential that repels candidates from reference features (orange) toward sparse regions; and (b) a diversity potential that enforces mutual repulsion among candidates to promote dispersion.}
    \vspace{-0.6cm}
    \label{fig9}
\end{figure}

Although the \textit{rarity score} enables global selection of sparse candidates, further optimization is needed to enhance local sparsity and inter-sample diversity. Inspired by physical potential fields, we model candidate samples as repulsive particles in the feature space and define a density-driven potential function that quantifies local feature crowding. Specifically, each candidate is guided to reside in a low-potential (i.e., sparsely populated) region by explicitly minimizing a smooth neighborhood density measure. 

Formally, for each selected candidate feature $\tilde{f}$, we define its local neighborhood as the reference samples within an $\epsilon$-radius in the semantic space, where the candidate serves as the center and neighbors are same-class reference samples. To enable gradient-based optimization, the hard neighbor count is replaced with a sigmoid-based smooth approximation $\sigma$:
\begin{equation}
N_i = \sum_{j}\sigma(\epsilon-d_{ij}),
\end{equation}
where $\epsilon$ defines the effective neighborhood radius, and
\begin{equation}
d_{ij} = || \tilde{f}_i - f_j ||, \tilde{f}_i \in \tilde{C}_k, f_j \in R_k.
\end{equation}

We then formulate a sparsity potential function that penalizes high local density:
\begin{equation}
\Phi_{sparsity} = \frac{1}{B}\sum_{i=1}^{B}{N_i}^{2}.
\label{eqsp}
\end{equation}
By minimizing this potential through gradient-based updates to the candidate feature $\tilde{f}$, the optimized features are adaptively repelled from high-density clusters and pushed into sparse regions of the semantic space.

Meanwhile, to further ensure that the generated candidates themselves remain sufficiently diverse and do not collapse into redundant clusters, we introduce an additional repulsion mechanism among the candidate samples. Similarly, we enforce pairwise separation between candidate features by defining a diversity potential:
\begin{equation}
\Phi_{diversity} = \frac{1}{B}\sum_{i=1}^{B}{\sum_{j\neq i}\sigma(\epsilon-d'_{ij})}^{2},
\label{eqdiv}
\end{equation}
where
\begin{equation}
d'_{ij} = || \tilde{f}_i - \tilde{f}_j ||, \tilde{f}_i, \tilde{f}_j \in \tilde{C}_k.
\end{equation}

Minimizing $\Phi_{diversity}$ encourages candidate samples to spread apart within the semantic space, discouraging trivial duplication and enforcing local intra-class diversity.

The final objective thus combines density-based repulsion from training samples and mutual repulsion among candidates, yielding a balanced potential field:
\begin{equation}
\Phi = \Phi_{sparsity} + \Phi_{diversity}.
\end{equation}

This dual-potential approach provides targeted control over how generated samples align with underrepresented regions and maintain sufficient pairwise separation, resulting in optimized features that are both sparse and diverse.

Before feeding the optimized sparse features into the diffusion model, we filter them based on confidence scores from the task-aware classifier to ensure the generation quality, retaining only those above a predefined lower threshold. To maintain balanced class coverage, low-confidence samples are replaced with high-fidelity GMSS samples drawn from $\mathcal{N}(\mu_k, 0.5\Sigma_k)$. This yields a final condition set that combines sparsity, diversity, and faithfulness for more effective diffusion-based augmentation. To ensure efficiency and consistency, all condition features are synthesized offline, organized by class, before initiating the reverse diffusion process. The overall algorithm is provided in Algorithm~1 in the Supplementary Material.\footnote{Supplementary materials: https://github.com/Lchenuu/SASG-DA.}

\begin{table}[!t]
    \caption{Characteristics and setup of three public sEMG datasets.} 
    \label{tab1}
    \centering
    \small
    \vspace{-0.1cm}
    \setlength{\tabcolsep}{0.95mm}{
        \begin{tabular}{lcccc}
        \toprule
        Dataset &  Ninapro DB2 & Ninapro DB4 & NinaproDB7 \\
        \midrule
        Subjects  & 40 & 10 & 20  \\
        Channels  & 12 & 12 & 12   \\
        Sampling rate  & 2000Hz & 2000Hz & 2000Hz \\
        Trials  & 6 & 6 & 6  \\
        Training Trials & 1,3,4,6 & 1,3,4,6 & 1,3,4,6 \\
        Testing Trials  & 2,5 & 2,5 & 2,5 \\
        Gestures  & 49 & 52 & 40   \\
        \bottomrule
        \end{tabular}}
    \vspace{-0.6cm}
\end{table}

\section{Experiments and results}

\subsection{Datasets}
To evaluate the effectiveness of the proposed approach, we conduct data augmentation experiments on three public sEMG benchmark datasets, namely Ninapro DB2~\cite{atzori2014electromyography}, DB4~\cite{pizzolato2017comparison}, and DB7~\cite{krasoulis2017improved}, as summarized in Table~\ref{tab1}. These datasets encompass a diverse range of gesture classes and subjects, providing a comprehensive testbed for evaluating generalization. During preprocessing, the raw sEMG signals of each subject are segmented via a sliding window of length 200 ms with a step size of 40 ms, and then standardized channel-wise. To ensure consistency with previous studies~\cite{wang2024transformer, al2024tcnn} and focus on gesture-related categories, the rest class is excluded during preprocessing. Following established protocols~\cite{wang2024transformer, al2024tcnn} for gesture recognition based on sEMG, we adopt a cross-trial paradigm in which the training and testing sets are divided based on the trial information, as mentioned in Table~\ref{tab1}.

\begin{table*}[!t]
\centering
\caption{Performance comparison with SOTAs in terms of ACC (\%), Pre(\%), Rec(\%) and F1(\%) on dataset DB7 using three different backbones. The results report the mean and standard deviation across subjects. Best performances are highlighted in bold.}
\label{tab2}
\resizebox{\textwidth}{!}{%
\begin{tabular}{l *{12}{S[table-format=2.2(2)]}}
\toprule
\multicolumn{1}{c}{\multirow{2}{*}{Aug. Method}}  & \multicolumn{4}{c}{Crossformer} & \multicolumn{4}{c}{TDCT} & \multicolumn{4}{c}{STCNet} \\ 
\cmidrule(lr){2-5} \cmidrule(lr){6-9} \cmidrule(lr){10-13} 
\multicolumn{1}{c}{} & \multicolumn{1}{c}{Acc} & \multicolumn{1}{c}{Pre} & \multicolumn{1}{c}{Rec}  & \multicolumn{1}{c}{F1} & \multicolumn{1}{c}{Acc} & \multicolumn{1}{c}{Pre}  & \multicolumn{1}{c}{Rec} & \multicolumn{1}{c}{F1} & \multicolumn{1}{c}{Acc} & \multicolumn{1}{c}{Pre}  & \multicolumn{1}{c}{Rec} & \multicolumn{1}{c}{F1} \\
    \midrule
Baseline  & 77.97\pm4.07$^{***}$ & 78.86\pm3.63 & 78.37\pm3.78 & 78.61\pm3.74 & 73.54\pm5.28$^{***}$  & 74.99\pm3.78 & 73.96\pm4.76 & 74.46\pm4.55 & 73.61\pm5.50$^{***}$ & 75.10\pm4.67 & 74.17\pm4.90 & 74.63\pm4.86  \\
Jittering \& Scaling  & 78.05\pm4.82$^{***}$ & 79.12\pm4.30 & 78.44\pm4.45  & 78.78\pm4.40  & 74.10\pm5.17$^{***}$ & 75.44\pm3.94 & 74.53\pm4.60 & 74.92\pm4.44 & 72.28\pm5.87$^{***}$ & 73.98\pm4.85 & 72.88\pm5.24 & 73.43\pm5.15  \\
Upsample & \hspace{1mm}\underline{80.71$\pm$3.76}$^{*}$ &  \underline{81.42$\pm$3.30} & \underline{81.17$\pm$3.26} &  \underline{81.29$\pm$3.32}  & 74.67\pm5.19$^{***}$ & 75.95\pm3.82 & 75.00\pm4.67 & 75.47\pm4.50 & 79.92\pm5.51$^{**}$ & \underline{81.64$\pm$4.41} & 80.50\pm4.84 & 81.07\pm5.07 \\
FreqMask  & 77.99\pm4.90$^{***}$ & 79.02\pm4.32 & 78.62\pm4.38 & 78.82\pm4.47  & 75.59\pm5.85$^{***}$ & 77.11\pm4.04 & 76.10\pm5.38 & 76.60\pm5.34  & 78.87\pm4.43$^{***}$ & 80.06\pm3.76 & 79.47\pm3.87 & 79.76\pm3.97 \\
RIM  & 79.04\pm3.97$^{***}$ & 79.75\pm3.58 & 79.49\pm3.66 & 79.62\pm3.67  & 76.12\pm5.56$^{***}$ & 77.28\pm4.17 & 76.55\pm5.12 & 76.91\pm5.04 & 76.93\pm5.25$^{***}$ & 78.18\pm4.63 & 77.48\pm4.71 & 77.83\pm4.72 \\
Dominant Shuffle & 79.41\pm4.47$^{***}$ & 80.22\pm3.93 & 79.80\pm4.13 & 80.01\pm4.08  & 75.21\pm5.90$^{***}$ & 77.03\pm4.15 & 75.67\pm5.59 & 76.34\pm5.46 & 78.92\pm4.46$^{***}$ & 80.37\pm3.74 & 79.46\pm4.00 & 79.91\pm4.00\\
\midrule
Mixup & 76.99\pm5.01$^{***}$ & 77.79\pm4.55 & 77.58\pm4.53 & 77.68\pm4.59 & 75.17\pm4.33$^{***}$ & 76.03\pm3.88 & 75.68\pm3.96 & 75.85\pm3.94 & 73.90\pm5.72$^{***}$ & 75.18\pm5.02 & 74.55\pm5.20 & 74.86\pm5.17  \\
STAug  & 77.43\pm5.15$^{***}$ & 78.30\pm4.60 & 77.95\pm4.66 & 78.12\pm4.67  & 75.32\pm4.32$^{***}$ & 76.29\pm3.69 & 75.72\pm3.99 & 76.00\pm3.91 & 73.53\pm5.65$^{***}$ & 74.77\pm5.03 & 74.22\pm5.01 & 74.49\pm5.06 \\
FreqMix & 79.79\pm4.11$^{***}$ & 80.96\pm3.54 & 80.22\pm3.76 & 80.59\pm3.72 & 76.24\pm3.61$^{*}$ & \underline{77.68$\pm$3.04} & 76.56\pm3.34 & \underline{77.12$\pm$3.23} & 78.35\pm4.42$^{***}$ & 79.94\pm3.59 & 78.86\pm3.96 & 79.40\pm3.99 \\
SimPSI  & 78.23\pm4.31$^{***}$ & 78.94\pm3.89 & 78.60\pm4.00 & 78.77\pm3.98  & 74.12\pm5.33$^{***}$ & 75.43\pm3.81 & 74.52\pm4.78 & 74.97\pm4.69 & 74.51\pm5.36$^{***}$ & 75.92\pm4.58 & 75.16\pm4.77 &  75.54\pm4.70 \\
\midrule
E-TRGAN & 78.35\pm4.72$^{***}$ & 78.95\pm3.89 & 78.83\pm4.56
& 78.89\pm4.03 & 74.16\pm3.77$^{***}$ & 76.02\pm4.91 & 75.85\pm3.68
& 75.93\pm4.44 & 74.24\pm3.55$^{***}$ & 75.30\pm4.21 & 74.95\pm3.12
& 75.12\pm4.87 \\
DiffMix  & 78.14\pm4.67$^{***}$ & 78.90\pm4.15 & 78.52\pm4.31 & 78.71\pm4.26  & 74.44\pm4.74$^{***}$ & 75.40\pm4.01 & 74.86\pm4.30 & 75.13\pm4.19 & 74.26\pm5.31$^{***}$ & 75.51\pm4.60 & 74.89\pm4.69 & 75.20\pm4.69 \\
SSE-LDM & 78.13\pm4.61$^{***}$ & 78.86\pm4.11 & 78.53\pm4.23 & 78.70\pm4.19 & 74.64\pm4.54$^{***}$ & 75.55\pm4.05 & 75.03\pm4.20 & 75.29\pm4.16 & 73.69\pm5.53$^{***}$ & 74.94\pm4.96 & 74.24\pm5.06 & 74.59\pm5.05 \\
EEG-Enhancing & 78.40\pm4.36$^{***}$ & 79.27\pm3.87 & 78.80\pm4.06 & 79.04\pm3.96 & 73.84\pm6.87$^{***}$ & 75.41\pm4.65 & 74.23\pm6.35 & 74.79\pm5.54 & 73.61\pm5.41$^{***}$ & 75.10\pm4.47 & 74.24\pm4.73 & 74.67\pm4.60 \\ 
NTD & 78.71\pm3.59$^{***}$ & 79.58\pm3.22 & 79.08\pm3.31 & 79.33\pm3.26 & 75.86\pm4.64$^{***}$ & 77.18\pm3.79 & 76.23\pm4.26 & 76.70\pm4.02 & 77.82\pm4.93$^{***}$ & 79.03\pm4.32 & 78.30\pm4.52 & 78.66\pm4.42 \\
PatchEMG & 79.37\pm3.80$^{***}$ & 80.12\pm3.39 & 79.87\pm3.43 & 79.99\pm3.46 & 76.29\pm4.24$^{**}$ & 77.34\pm3.57 & \underline{76.78$\pm$3.73} & 77.06\pm3.72 & 79.14\pm4.95$^{***}$ & 80.29\pm4.22 & 79.74\pm4.32 & 80.01\pm4.36 \\
DESAM & 79.73\pm4.20$^{***}$ & 80.58\pm3.70 & 80.19\pm3.84 & 80.38\pm3.79 & 75.88\pm4.81$^{**}$ & 77.20\pm3.90 & 76.18\pm4.38 & 76.69\pm4.23 & 78.64\pm5.09$^{***}$ & 79.61\pm4.55 & 79.04\pm4.70 & 79.32\pm4.67 \\
Minority & 79.86\pm4.02$^{***}$ & 80.30\pm3.69 & 80.37\pm3.68 & 80.33\pm3.76 & 75.14\pm4.40$^{***}$ & 75.86\pm4.08 & 75.67\pm3.98 & 75.76\pm4.15 & 79.84\pm4.37$^{***}$ & 80.53\pm3.89 & 80.37\pm3.87 & 80.45\pm3.94  \\
CADS & 80.41\pm4.11$^{***}$ & 80.83\pm3.83 & 80.86\pm3.75 & 80.84\pm3.86  & \hspace{4mm}\underline{76.36$\pm$5.17}$^{***}$ & 77.16\pm4.38 & 76.75\pm4.62 & 76.95\pm4.83 & \hspace{4mm}\underline{80.55$\pm$4.57}$^{***}$ & 81.28\pm4.08 & \underline{81.07$\pm$4.04} & \underline{81.17$\pm
$4.09} \\
\midrule
SASG-DA (Ours)  & \textbf{81.31$\pm$3.99} & \textbf{81.89$\pm$3.64} & \textbf{81.75$\pm$3.64} & \textbf{81.82$\pm$3.70} & \textbf{78.77$\pm$4.85} & \textbf{79.73$\pm$3.99}  & \textbf{79.23$\pm$4.58} & \textbf{79.48$\pm$4.59} & \textbf{82.15$\pm$4.59}  & \textbf{83.07$\pm$3.98} & \textbf{82.68$\pm$4.14} & \textbf{82.87$\pm$4.13}  \\
    \bottomrule
\end{tabular}}
\begin{tablenotes}
\footnotesize
\item $^{*}$ $p<0.05$, $^{**}$ $p<0.01$, $^{***}$ $p<0.001$, $N=20$.
\end{tablenotes}
\vspace{-0.6cm}
\end{table*}

\begin{table*}[!t]
\centering
\caption{Performance comparison with SOTAs in terms of ACC (\%), Pre(\%), Rec(\%) and F1(\%) on dataset DB4 using three different backbones. The results report the mean and standard deviation across subjects. Best performances are highlighted in bold.}
\label{tab3}
\resizebox{\textwidth}{!}{%
\begin{tabular}{l *{12}{S[table-format=2.2(2)]}}
\toprule
\multicolumn{1}{c}{\multirow{2}{*}{Aug. Method}}  & \multicolumn{4}{c}{Crossformer} & \multicolumn{4}{c}{TDCT} & \multicolumn{4}{c}{STCNet} \\ 
\cmidrule(lr){2-5} \cmidrule(lr){6-9} \cmidrule(lr){10-13} 
\multicolumn{1}{c}{} & {Acc} & {Pre} & {Rec}  & {F1} & {Acc} & {Pre}  & {Rec} & {F1} & {Acc} & {Pre}  & {Rec} & {F1} \\
    \midrule
Baseline  & 65.78\pm5.88$^*$ & 66.97\pm5.95 & 67.49\pm6.15 & 67.23\pm6.12 &
62.62\pm4.97$^*$ & 64.18\pm5.34 & 64.34\pm5.18 & 64.26\pm5.31 &
61.25\pm7.03$^*$ & 63.43\pm7.04 & 63.05\pm7.43 & 63.24\pm7.39  \\
Jittering \& Scaling  & 66.01\pm5.75$^*$ & 67.35\pm5.85 & 67.80\pm5.98 & 67.57\pm5.99 &
63.05\pm5.02$^*$ & 64.60\pm5.28 & 64.75\pm5.20 & 64.67\pm5.30 &
59.97\pm6.89$^*$ & 62.23\pm7.01 & 61.74\pm7.22 & 61.98\pm7.22 \\
Upsample & 67.62\pm6.02$^*$ & 69.23\pm6.28 & 69.29\pm6.28 & 69.26\pm6.34 &
63.37\pm4.73$^*$ & 65.16\pm5.21 & 65.23\pm4.96 & 65.19\pm5.05 &
\underline{68.44$\pm$5.90} & \underline{70.40$\pm$6.05} & \underline{70.14$\pm$6.20} & \underline{70.27$\pm$6.25} \\
FreqMask  & 66.67\pm6.16$^*$ & 68.08\pm6.22 & 68.35\pm6.23 & 68.21\pm6.37 &
64.50\pm5.55$^*$ & 66.31\pm5.76 & 66.23\pm5.61 & 66.27\pm5.80 &
67.08\pm6.25$^*$ & 68.99\pm6.29 & 68.82\pm6.33 & 68.90\pm6.51 \\
RIM & 67.18\pm5.88$^*$ & 68.31\pm6.06 & 68.90\pm6.12 & 68.60\pm6.19 &
64.47\pm5.19$^*$ & 65.75\pm5.39 & 66.21\pm5.38 & 65.98\pm5.48 &
65.08\pm6.44$^*$ & 66.58\pm6.33 & 66.77\pm6.67 & 66.67\pm6.66\\
Dominant Shuffle & 67.21\pm5.77$^*$ & 68.41\pm5.84 & 68.96\pm5.98 & 68.68\pm6.00 &
64.22\pm5.59$^*$ & 66.01\pm5.63 & 66.06\pm5.70 & 66.03\pm5.84 &
65.89\pm6.58$^*$ & 67.89\pm6.34 & 67.69\pm6.74 & 67.79\pm6.80 \\
\midrule
Mixup & 66.21\pm5.68$^*$ & 67.48\pm5.67 & 67.86\pm5.81 & 67.67\pm5.88 &
64.50\pm4.81$^*$ & 65.96\pm5.04 & 66.20\pm5.00 & 66.08\pm5.14 &
62.13\pm6.84$^*$ & 63.95\pm6.94 & 63.82\pm7.06 & 63.88\pm7.14 \\
STAug  & 66.41\pm5.89$^*$ & 67.77\pm6.03 & 68.09\pm6.06 & 67.93\pm6.14 &
63.96\pm5.70$^*$ & 65.31\pm5.88 & 65.67\pm5.82 & 65.49\pm5.95 &
61.46\pm6.95$^*$ & 63.36\pm7.12 & 63.10\pm7.25 & 63.23\pm7.20 \\
FreqMix & 68.51\pm5.72$^*$ & \underline{70.03$\pm$5.78} & 70.25\pm5.84 & 70.14\pm5.94 &
65.45\pm5.15$^*$ & \underline{67.08$\pm$5.15} & 67.15\pm5.18 & 67.11\pm5.27 &
66.69\pm6.19$^*$ & 68.73\pm6.05 & 68.38\pm6.23 & 68.55\pm6.35 \\
SimPSI  & 66.72\pm6.02$^*$ & 67.93\pm5.99 & 68.40\pm6.10 & 68.16\pm6.19 &
64.06\pm5.43$^*$ & 65.54\pm5.58 & 65.78\pm5.58 & 65.66\pm5.73 &
62.44\pm7.51$^*$ & 64.48\pm7.38 & 64.06\pm7.85 & 64.27\pm7.79\\
\midrule
E-TRGAN &65.93\pm5.42$^*$ &67.70\pm6.11 &67.28\pm5.67 
&67.51\pm5.94 &63.11\pm5.33$^*$ &64.60\pm6.48 &64.55\pm5.58 &64.57\pm6.23 &62.66\pm5.21$^*$ 
&64.50\pm5.89 &64.54\pm5.07 &64.52\pm6.37 \\
DiffMix  & 66.08\pm6.07$^*$ & 67.21\pm6.09 & 67.81\pm6.29 & 67.51\pm6.31 &
62.14\pm5.64$^*$ & 63.35\pm5.88 & 63.83\pm5.89 & 63.59\pm5.96 &
63.22\pm6.45$^*$ & 64.89\pm6.37 & 64.99\pm6.85 & 64.94\pm6.74\\
SSE-LDM & 66.10\pm5.33$^*$ & 67.16\pm5.39 & 67.76\pm5.53 & 67.46\pm5.56 &
63.53\pm4.97$^*$ & 64.76\pm5.32 & 65.26\pm5.18 & 65.01\pm5.29 &
62.56\pm6.50$^*$ & 64.18\pm6.48 & 64.34\pm6.84 & 64.26\pm6.78  \\
EEG-Enhancing & 66.48\pm5.84$^{**}$ & 67.66\pm5.90 & 68.22\pm6.06 & 67.94\pm5.97 &
62.71\pm4.95$^{**}$ & 64.16\pm5.26 & 64.47\pm5.13 & 64.31\pm5.19 &
61.86\pm6.80$^{**}$ & 63.69\pm6.85 & 63.63\pm7.11 & 63.66\pm6.97 \\
NTD & 67.90\pm6.13$^{**}$ & 68.92\pm6.18 & 69.55\pm6.36 & 69.23\pm6.27 &
64.29\pm5.89$^{**}$ & 65.72\pm6.21 & 65.85\pm5.93 & 65.78\pm6.06 &
65.69\pm8.56$^{**}$ & 67.19\pm8.31 & 67.39\pm8.53 & 67.29\pm8.41 \\
PatchEMG & 67.50\pm5.09$^{**}$ & 68.93\pm5.41 & 69.26\pm5.49 & 69.09\pm5.44 &
64.68\pm4.85$^{**}$ & 66.38\pm5.24 & 66.37\pm5.02 & 66.37\pm5.14 &
67.45\pm6.13$^{**}$ & 69.42\pm6.09 & 69.21\pm6.37 & 69.31\pm6.33 \\ 
DESAM & 67.63\pm5.85$^{**}$ & 68.82\pm5.97 & 69.33\pm6.15 & 69.07\pm6.16 &
63.91\pm5.42$^{**}$ & 65.24\pm5.68 & 65.60\pm5.56 & 65.42\pm5.69 &
64.06\pm7.32$^{**}$ & 65.67\pm7.01 & 65.90\pm7.53 & 65.78\pm7.41  \\
Minority  & 68.39\pm5.94$^*$ & 69.49\pm5.92 & 70.03\pm6.18 & 69.76\pm6.17 &
64.47\pm5.78$^*$ & 65.86\pm5.85 & 66.13\pm5.97 & 65.99\pm6.00 &
66.74\pm7.20$^*$ & 68.38\pm7.21 & 68.42\pm7.45 & 66.40\pm7.56\\
CADS  & \hspace{1mm}\underline{68.91$\pm$5.60$^*$} & 69.99\pm5.61 & \underline{70.57$\pm$5.94} & \underline{70.28$\pm$5.87} &
\underline{65.58$\pm$5.06} & 67.03\pm5.09 & \underline{67.33$\pm$5.27} & \underline{67.18$\pm$5.23} &
67.45\pm6.82$^*$ & 68.84\pm6.96 & 69.13\pm7.13 & 68.98\pm7.20 \\
\midrule
SASG-DA (Ours)  & \textbf{69.73$\pm$5.79} & \textbf{70.73$\pm$5.79} & \textbf{71.38$\pm$5.99} & \textbf{71.05$\pm$6.02} &
\textbf{67.74$\pm$5.72} & \textbf{69.02$\pm$5.86} & \textbf{69.42$\pm$5.88} & \textbf{69.22$\pm$5.97} &
\textbf{70.05$\pm$6.26} & \textbf{71.37$\pm$6.27} & \textbf{71.75$\pm$6.42} & \textbf{71.56$\pm$6.53}  \\
    \bottomrule
\end{tabular}}
\begin{tablenotes}
\footnotesize
\item $^{*}$ $p<0.05$, $^{**}$ $p<0.01$, $^{***}$ $p<0.001$, $N=10$.
\end{tablenotes}
\vspace{-0.6cm}
\end{table*}

\begin{table*}[!t]
\centering
\caption{Performance comparison with SOTAs in terms of ACC (\%), Pre(\%), Rec(\%) and F1(\%) on dataset DB2 using three different backbones. The results report the mean and standard deviation across subjects. Best performances are highlighted in bold.}
\label{tab4}
\resizebox{\textwidth}{!}{%
\begin{tabular}{l *{12}{S[table-format=2.2(2)]}}
\toprule
\multicolumn{1}{c}{\multirow{2}{*}{Aug. Method}}  & \multicolumn{4}{c}{Crossformer} & \multicolumn{4}{c}{TDCT} & \multicolumn{4}{c}{STCNet} \\ 
\cmidrule(lr){2-5} \cmidrule(lr){6-9} \cmidrule(lr){10-13} 
\multicolumn{1}{c}{} & \multicolumn{1}{c}{Acc} & \multicolumn{1}{c}{Pre} & \multicolumn{1}{c}{Rec}  & \multicolumn{1}{c}{F1} & \multicolumn{1}{c}{Acc} & \multicolumn{1}{c}{Pre}  & \multicolumn{1}{c}{Rec} & \multicolumn{1}{c}{F1} & \multicolumn{1}{c}{Acc} & \multicolumn{1}{c}{Pre}  & \multicolumn{1}{c}{Rec} & \multicolumn{1}{c}{F1} \\
 \midrule
Baseline  & 74.63\pm6.10$^{***}$ & 75.89\pm5.68 & 75.44\pm5.55 & 75.66\pm5.67 & 70.00\pm6.19$^{***}$ & 71.44\pm5.60 & 70.83\pm5.58 & 71.13\pm5.65 & 69.71\pm6.61$^{***}$ & 71.86\pm5.97 & 70.83\pm5.93 & 71.34\pm6.03  \\
Jittering \& Scaling  & 74.36\pm6.34$^{***}$ & 75.75\pm5.84 & 75.20\pm5.78 & 75.47\pm5.88 & 70.08\pm6.34$^{***}$ & 71.67\pm5.68 & 70.94\pm5.67 & 71.30\pm5.75 & 68.46\pm7.03$^{***}$ & 70.78\pm6.32 & 69.61\pm6.30 & 70.19\pm6.37 \\
Upsample & 75.84\pm6.28$^{***}$ & 77.18\pm5.84 & 76.85\pm5.80 & 77.01\pm5.93 & 70.05\pm6.77$^{***}$ & 71.66\pm5.85 & 70.91\pm6.15 & 71.28\pm6.23 & \underline{74.38$\pm$9.15}  & \underline{76.69$\pm$7.07} & \underline{75.36$\pm$8.45} & \underline{76.02$\pm$8.58}\\
FreqMask  & 74.37\pm6.93$^{***}$ & 75.90\pm6.27 & 75.36\pm6.39 & 75.63\pm6.51 & 71.22\pm6.01$^{***}$ & 72.90\pm5.49 & 72.16\pm5.48 & 72.53\pm5.57 & 73.86\pm6.40$^{***}$ & 75.70\pm5.88 & 74.89\pm5.89 & 75.29\pm6.01 \\
RIM & 75.69\pm6.14$^{***}$ & 76.81\pm5.75 & 76.53\pm5.61 & 76.67\pm5.75 & 71.36\pm6.34$^{***}$ & 72.72\pm5.74 & 72.28\pm5.73 & 72.50\pm5.82 & 72.14\pm6.87$^{***}$ & 73.78\pm6.21 & 73.20\pm6.21 & 73.49\pm6.29\\
Dominant Shuffle & 75.82\pm6.10$^{***}$ & 77.10\pm5.69 & 76.63\pm5.62 & 76.86\pm5.73 & 71.02\pm6.58$^{***}$ & 73.01\pm5.59 & 71.91\pm5.91 & 72.46\pm6.01 & 74.38\pm6.72$^{***}$ & 76.17\pm6.03 & 75.33\pm6.04 & 75.75\pm6.11 \\
\midrule
Mixup & 74.45\pm6.28$^{***}$ & 75.65\pm5.81 & 75.45\pm5.70 & 75.55\pm5.82 & 71.13\pm5.72$^{***}$ & 72.47\pm5.19 & 72.12\pm5.15 & 72.29\pm5.25 & 69.46\pm6.84$^{***}$ & 71.44\pm6.12 & 70.58\pm6.36 & 71.01\pm6.42 \\
STAug  & 74.49\pm6.44$^{***}$ & 75.65\pm5.99 & 75.40\pm5.87 & 75.52\pm5.99 & 71.02\pm5.92$^{***}$ & 72.36\pm5.47 & 71.94\pm5.31 & 72.15\pm5.43 & 69.52\pm7.11$^{***}$ & 71.45\pm6.53 & 70.72\pm6.54 & 71.08\pm6.60\\
FreqMix & \hspace{1mm}\underline{77.07$\pm$5.72}$^{*}$ & \underline{78.57$\pm$5.25} & 77.86\pm5.22 & \underline{78.21$\pm$5.34} & 71.47\pm6.53$^{***}$ & \underline{73.56$\pm$5.52} & 72.33\pm5.91 & 72.94\pm6.10 & 73.92\pm6.91$^{***}$ & 75.97\pm5.92 & 74.94\pm6.24 & 75.45\pm6.42 \\
SimPSI  & 74.58\pm6.06$^{***}$ & 75.74\pm5.68 & 75.42\pm5.54 & 75.58\pm5.67 & 70.52\pm5.93$^{***}$ & 71.98\pm5.41 & 71.43\pm5.28 & 71.70\pm5.39 & 70.91\pm7.09$^{***}$ & 72.80\pm6.43 & 72.00\pm6.48 & 72.40\pm6.56 \\
\midrule
E-TRGAN &75.17\pm5.42$^{***}$ &75.85\pm6.33 &75.90\pm5.96 &75.87\pm6.12 &70.41\pm5.28$^{***}$ &72.05\pm6.41 &71.90\pm5.73 &71.98\pm6.25 &70.21\pm5.15$^{***}$ &71.70\pm5.89 &71.82\pm5.07 &71.76\pm6.48 \\
DiffMix  & 74.84\pm6.28$^{***}$ & 76.07\pm5.84 & 75.74\pm5.71 & 75.90\pm5.86 & 69.47\pm6.61$^{***}$ & 70.86\pm6.08 & 70.49\pm5.98 & 70.67\pm6.07 & 70.33\pm6.55$^{***}$ & 72.01\pm6.11 & 71.49\pm5.97 & 71.75\pm6.10 \\
SSE-LDM & 74.73\pm6.22$^{***}$ & 75.88\pm5.83 & 75.61\pm5.68 & 75.74\pm5.81 & 70.28\pm6.45$^{***}$ & 71.62\pm5.97 & 71.26\pm5.84 & 71.44\pm5.95 & 69.91\pm7.01$^{***}$ & 71.62\pm6.41 & 71.08\pm6.41 & 71.34\pm6.51 \\
EEG-Enhancing & 75.10\pm5.99$^{***}$ & 76.35\pm5.51 & 75.89\pm5.45 & 76.12\pm5.47 & 69.93\pm6.33$^{***}$ & 71.45\pm5.79 & 70.84\pm5.70 & 71.15\pm5.74 & 70.08\pm7.05$^{***}$ & 72.11\pm6.27 & 71.22\pm6.32 & 71.66\pm6.29 \\
NTD & 75.76\pm5.86$^{***}$ & 77.14\pm5.42 & 76.59\pm5.31 & 76.86\pm5.36 & 70.87\pm6.83$^{***}$ & 72.68\pm6.01 & 71.77\pm6.23 & 72.22\pm6.11 & 72.00\pm7.26$^{***}$ & 73.94\pm6.45 & 73.02\pm6.56 & 73.47\pm6.50 \\
PatchEMG & 75.91\pm5.89$^{***}$ & 76.96\pm5.51 & 76.79\pm5.40 & 76.87\pm5.52 & 71.31\pm6.09$^{***}$ & 72.60\pm5.59 & 72.21\pm5.46 & 72.40\pm5.59 & 73.73\pm6.40$^{***}$ & 75.21\pm5.90 & 74.74\pm5.75 & 74.97\pm5.90 \\
DESAM & 76.32\pm6.25$^{***}$ & 77.53\pm5.82 & 77.16\pm5.72 & 77.34\pm5.84 & 70.47\pm6.41$^{***}$ & 72.21\pm5.75 & 71.32\pm5.76 & 71.76\pm5.81 & 72.55\pm7.04$^{***}$ & 74.41\pm6.39 & 73.49\pm6.40 & 73.94\pm6.41 \\
Minority  & 76.89\pm5.70$^{***}$ & 77.59\pm5.40 & 77.78\pm5.24 & 77.68\pm5.38 & 71.31\pm5.84$^{***}$ & 72.34\pm5.41 & 72.16\pm5.25 & 72.25\pm5.37 & 73.48\pm6.42$^{***}$ & 74.57\pm5.89 & 74.52\pm5.83 & 74.54\pm5.93 \\
CADS   & 77.03\pm5.67$^{***}$ & 77.74\pm5.36 & \underline{77.87$\pm$5.22} & 77.80\pm5.35 & \hspace{4mm}\underline{72.33$\pm$5.90}$^{***}$ & 73.27\pm5.57 & \underline{73.08$\pm$5.35} & \underline{73.17$\pm$5.53} & 74.33\pm6.36$^{***}$ & 75.34\pm5.93 & 75.29\pm5.84 & 75.31\pm5.96\\
\midrule
SASG-DA (Ours)  & \textbf{77.86$\pm$5.65} & \textbf{78.82$\pm$5.30} & \textbf{78.73$\pm$5.16} & \textbf{78.77$\pm$5.28} & \textbf{74.31$\pm$5.94} & \textbf{75.46$\pm$5.46} & \textbf{75.20$\pm$5.31} & \textbf{75.33$\pm$5.47} & \textbf{76.36$\pm$6.19}  & \textbf{77.59$\pm$5.70} & \textbf{77.29$\pm$5.60} & \textbf{77.44$\pm$5.72}    \\
 \bottomrule
\end{tabular}}
\begin{tablenotes}
\footnotesize
\item $^{*}$ $p<0.05$, $^{**}$ $p<0.01$, $^{***}$ $p<0.001$, $N=40$.
\end{tablenotes}
\vspace{-0.4cm}
\end{table*}

\subsection{Evaluation Metrics}
To comprehensively evaluate the effectiveness of our proposed SASG-DA approach for sEMG-based gesture recognition, we employ the following three categories of metrics: classification performance under the intra-subject paradigm, generation quality metrics, and sparsity evaluation metrics. All latent features involved in evaluation are extracted by a pretrained classifier.

We evaluate the classification performance under the intra-subject setting. Accuracy (ACC) is reported as the average classification accuracy across all subjects and serves as the primary indicator of the model's classification ability. In addition, precision (Pre), recall (Rec), and F1-score are calculated to offer a more detailed assessment of the model’s classification performance. To evaluate whether our method provides statistically meaningful improvements over existing augmentation strategies, we performed subject-wise Wilcoxon signed-rank tests on average accuracy across three random runs per subject for each backbone and applied Bonferroni correction to account for multiple comparisons.

To assess the quality of the generated samples, we adopt the Fréchet Inception Distance (FID)~\cite{heusel2017gans} and Category Accuracy Score (CAS).
FID quantifies the distributional similarity between real and generated samples in a latent feature space. 
CAS evaluates the semantic consistency of generated samples by measuring the classification accuracy of generated data using a classifier trained solely on real samples. It reflects the model’s ability to generate the intended target classes.

To further analyze the diversity and sparsity of generated samples, we employ three sparsity-aware metrics derived from~\cite{sehwag2022generating, um2024self}, following the same evaluation settings: (1) Average KNN Distance (AvgKNN) measures the average distance from a sample to its nearest neighbors in the feature space ($k=5$). (2) Local Outlier Factor (LOF) assesses how the local density of a sample deviates from that of its neighbors ($k=20$). (3) \textit{Rarity score}~\cite{han2024rarity} explicitly quantifies the rarity of individual generated samples ($k=5$). We compare the sparsity of generated samples against that of real samples.

\subsection{Experimental Settings}
\label{settings}

The experiments consist of diffusion-based data augmentation and downstream classification. The diffusion model is built on a 1D U-Net~\cite{aristimunha2023synthetic} and trained only on the original training set with 1000 diffusion steps (cosine noise schedule). It uses the $x_0$-prediction objective and Adam (lr=1e-5, batch size=128) for 20,000 iterations. A conditional dropout rate of 0.05 is applied. During sampling, DDIM with 500 steps is used. The involved pretrained classifier (Crossformer) during the generation is trained on the original training set for 100 epochs using SGD (lr=0.01), and 128-dimensional semantic representations are extracted from the last fully connected layer. For SASS, hyperparameters are set to 200 iterations (\textit{iter}), a neighborhood radius ($\epsilon$) of 3.0, a learning rate ($\eta$ ) of 10.0, and a confidence threshold of 0.15.

Following~\cite{wang2024enhance}, original training and generated samples are shuffled to form an augmented dataset twice the original size, used only for training while test sets remain fully real. To ensure class balance, the synthetic data followed the same class distribution as the training set. Downstream classifiers are trained with SGD (initial lr=0.01, decayed by 0.1 at epoch 60), batch size 256, for 100 epochs. Results are averaged over three runs with different random seeds. All experiments are conducted in PyTorch 2.3.0 on an NVIDIA RTX 4090 GPU.

To evaluate the data augmentation performance of our proposed approach, we adopt three backbone models: Crossformer~\cite{zhang2023crossformer}, TDCT~\cite{wang2024transformer}, and STCNet~\cite{yang2025stcnet}. More details about these backbone architectures and the corresponding parameter settings are provided in the Supplementary Material.

\subsection{Comparison with the State-of-the-arts}
We compare our method against a comprehensive set of state-of-the-art data augmentation approaches. Specifically, we evaluate five single-sample DA methods tailored for time series: Jittering \& Scaling~\cite{um2017data}, Upsample~\cite{semenoglou2023data}, FreqMask~\cite{chen2023fraug}, RIM~\cite{aboussalah2023recursive}, and Dominant Shuffle~\cite{zhao2024dominant}. We also consider four mix-based DA methods: Mixup~\cite{zhang2018mixup}, STAug~\cite{zhang2023towards}, FreqMix~\cite{chen2023fraug}, and SimPSI~\cite{ryu2024simpsi}. In addition, diffusion-based DA methods are included. DiffMix~\cite{wang2024enhance} is a simple yet effective generative data augmentation strategy that leverages inter-class interpolation to produce diverse samples. Minority~\cite{um2024self} leverages gradient guidance to synthesize samples in low-density regions. CADS~\cite{sadat2024cads} enhances sample diversity by injecting stochastic perturbations into conditional inputs during the inference process. PatchEMG~\cite{xiong2024patchemg} adopts a patch-based training strategy, and during inference, it generates complete-length signals while employing classifier-free guidance (CFG) to ensure generation quality. SSE-LDM~\cite{aristimunha2023synthetic}, EEG-Enhancing~\cite{siddhad2024enhancing}, NTD~\cite{vetter2024generating}, and DESAM~\cite{luo2025diffusion} are diffusion-based models for EEG signal generation. Among them, DESAM constructs an imputation-based diffusion process and temporal mixup with real samples to improve generation faithfulness. We also include E-TRGAN~\cite{zhao2024trgan} for comparison, which applies a Transformer-based GAN for sEMG signal reconstruction. For single-sample and mix-based DA methods, we adopt a batch-wise augmentation strategy, whereas for generative-based methods and RIM, we jointly train the original and augmented samples to preserve data distribution better and improve model generalization. For fairness, all SOTA methods are reproduced using official code when available, or reimplemented based on recommended or optimal settings.

The results in Table~\ref{tab2}, Table~\ref{tab3}, and Table~\ref{tab4} demonstrate that our method consistently outperforms the state-of-the-art (SOTA) on three datasets and with all three backbones, which confirms its generalizability and consistent effectiveness when used as a data augmentation method. Moreover, at the subject level, our method shows statistically significant improvements ($p<0.05$) over nearly all SOTA approaches, with relatively small standard deviations across three runs (0.02–0.11), indicating robust performance (see in the Supplementary Material). Compared to the next best method, CADS~\cite{sadat2024cads}, which enhances diversity in diffusion-based augmentation, our method achieves an average accuracy gain of \textbf{1.7}\% across all datasets and backbones. This confirms the effectiveness of guiding sample generation toward sparse regions, resulting in more informative and complementary samples for data augmentation. Compared to Minority~\cite{um2024self}, our approach achieves better performance by not only improving diversity but also enhancing the faithfulness, i.e., their alignment with class-consistent semantics. This improvement in sample faithfulness allows the augmented data to better support classifier learning, leading to more substantial gains in accuracy.

\begin{table}[!b]
\vspace{-0.6cm}
\centering
\caption{Ablations of each module in terms of ACC (\%) on DB7 and DB4. Best performances are highlighted in bold.}
\label{tab5}
\resizebox{\columnwidth}{!}{%
\begin{tabular}{cccccccccc}
\toprule
\multirow{2}{*}{Method} &\multicolumn{3}{c}{DB7} & \multicolumn{3}{c}{DB4} \\
\cmidrule(lr){2-4} \cmidrule(lr){5-7}
& Crossformer & TDCT & STCNet  &Crossformer & TDCT & STCNet  \\
\midrule
Baseline & 77.97 & 73.54 & 73.61 & 65.78 & 62.62 & 61.25  \\
Label Condition & 80.49 & 77.09 & 80.62 & 68.81 & 66.04 & 68.19  \\
GMSS & 80.87  & 77.57 & 81.67 & 69.44 & 66.99 & 69.41 \\
SASG-DA & \textbf{81.31} & \textbf{78.77} & \textbf{82.15} & \textbf{69.73} & \textbf{67.74} & \textbf{70.05}  \\
\bottomrule
\end{tabular}}
\vspace{-0.2cm}
\end{table}

\begin{figure}[!b]
    \centering
    \vspace{-0.5cm}
    \subfigure[]{
        \includegraphics[width=0.4\textwidth]{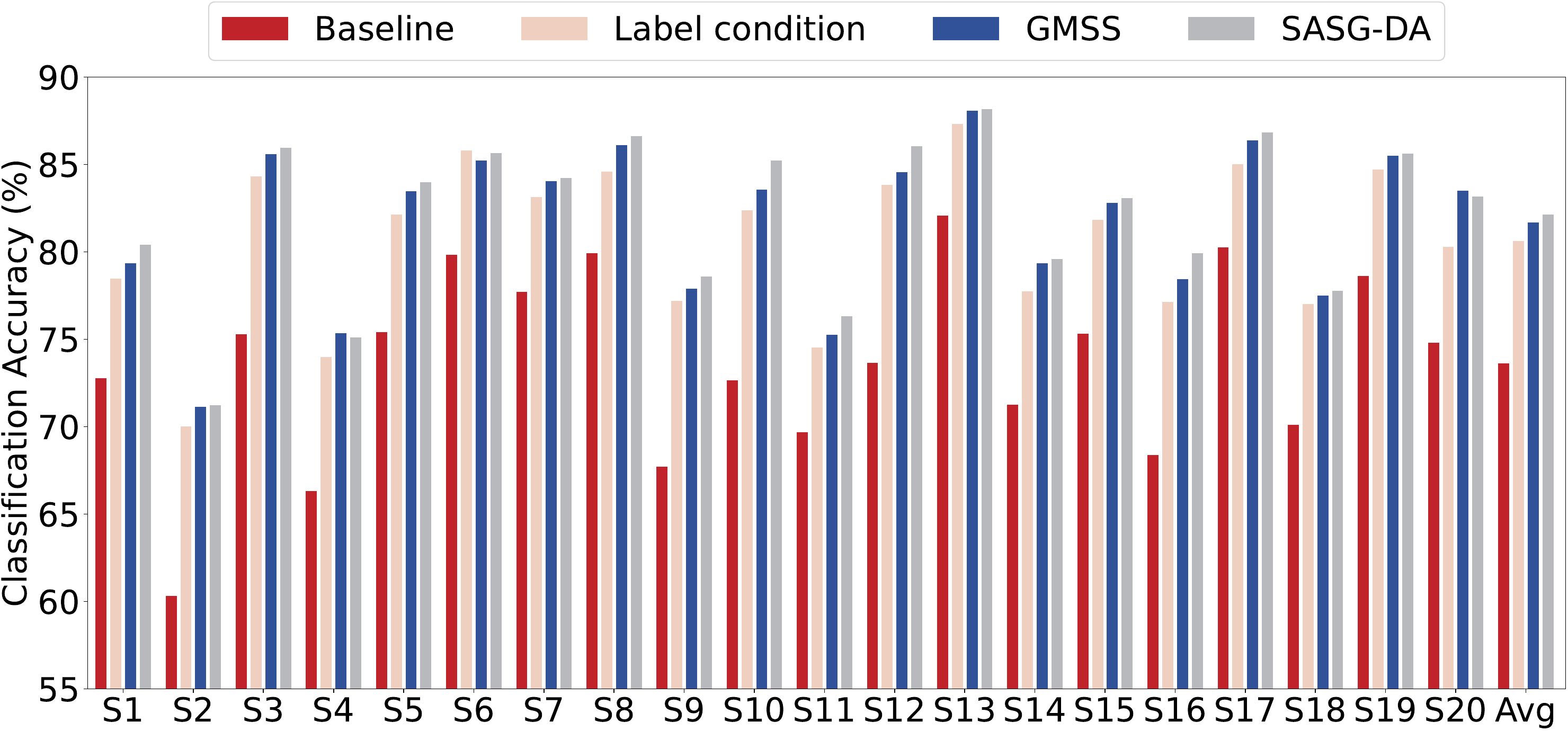}
    }

    \vspace{-0.2cm}
   \subfigure[]{
        \centering
        \includegraphics[width=0.4\textwidth]{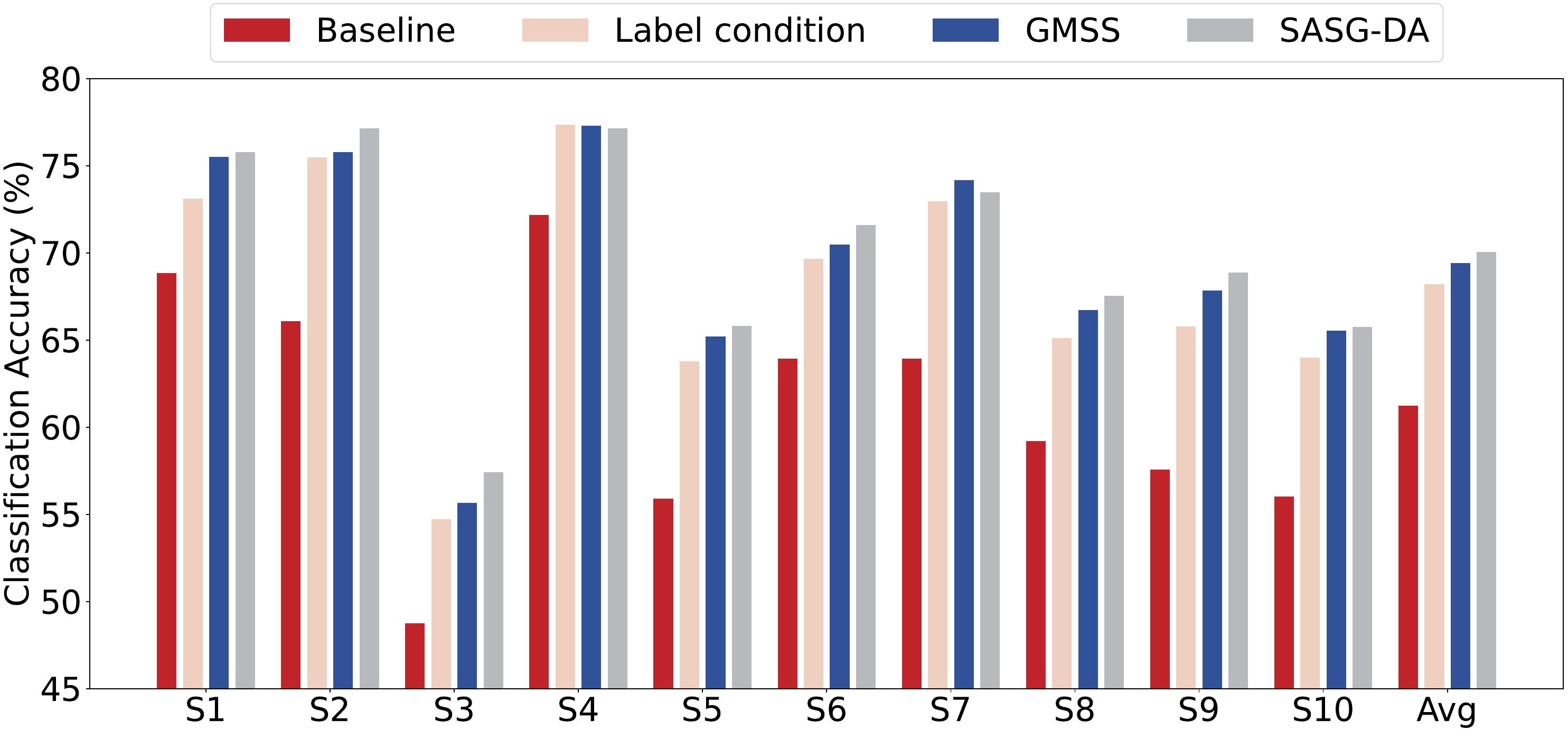}
    }
     \vspace{-0.2cm}
    \caption{Classification accuracy (\%) ablation results for all the subjects from (a) DB7 and (b) DB4.}
    \label{sub_ablation}
\end{figure}

\begin{table*}[thbp]
\centering
\caption{Hyperparameters and Semantic Encoder ablations on DB7 and DB4 under different settings for three backbones.}
\small
\renewcommand{\arraystretch}{0.85}
\resizebox{0.88\textwidth}{!}{%
\begin{tabular}{cccc|ccccc|ccccc}
\toprule
\multirow{2}{*}{\textit{iter}} & \multirow{2}{*}{$\epsilon$} & \multirow{2}{*}{k} & \multirow{2}{*}{threshold} 
& \multicolumn{5}{c}{DB7} 
& \multicolumn{5}{|c}{DB4}  \\
\cmidrule{5-9} \cmidrule{10-14}
 & &  & &  FID & CAS & Crossformer & TDCT & STCNet 
& FID & CAS & Crossformer & TDCT & STCNet\\
\midrule
200 & 2.0 & 50 & 0.15 & 1.32 & 61.08 & 81.07 & 78.51 & 81.82  & 1.39 & 57.46 & 69.51 & 67.84 & 69.98  \\
200 & 2.5 & 50 & 0.15 &1.33 & 59.99 & 81.13 & 78.80 & 82.16 & 1.41 & 56.70 & 69.64 & 67.95 & 70.15 \\
200 & 3.0 & 50 & 0.15 &1.35 & 58.65& 81.31 & 78.77 & 82.15 & 1.42 & 56.28 & 69.73 & 67.74  & 70.05   \\
\midrule
100 & 3.0 & 50 & 0.15 & 1.32 & 59.83 & 80.99  & 78.92 & 82.02  &1.40 & 57.14 & 69.65 & 67.61 & 70.04  \\
200 & 3.0 & 50 & 0.15 &1.35 & 58.65 & 81.31 & 78.77 & 82.15 & 1.42 & 56.28 & 69.73 & 67.74 & 70.05    \\
300 & 3.0 & 50 & 0.15 & 1.37 & 57.62 & 81.24 & 78.83 & 82.18 & 1.43 & 55.65 & 69.59  & 67.78  & 70.08  \\

\midrule
200 & 3.0 & 10 & 0.15 & 1.26 & 60.92 & 80.94 & 78.71 & 82.06 & 1.22 & 58.92 & 69.41 & 68.02 & 69.94 \\
200 & 3.0 & 30 & 0.15 & 1.31 & 60.13 & 80.86 & 78.80 & 82.07 & 1.36 & 58.10 & 69.54 & 67.69 & 69.75    \\
200 & 3.0 & 50  & 0.15 & 1.35 & 58.65 & 81.31 & 78.77 & 82.15 & 1.42 & 56.28 & 69.73 & 67.74 & 70.05    \\
\midrule
200 & 3.0  & 50& 0.05 & -&- & 81.00 & 78.63 & 81.93  & - & - & 69.72 & 67.59 & 69.86  \\
200 & 3.0  & 50& 0.15 & - & - & 81.31 & 78.77 & 82.15 &- &- & 69.73 & 67.74 & 70.05  \\
200 & 3.0  & 50 & 0.25 &- &- & 80.97 & 78.55 & 81.86 &- &- & 69.63  & 67.58 & 69.98  \\
\midrule
\multicolumn{4}{c|}{Crossformer} & 1.35 & 58.65 & 81.31 & 78.77 &82.15 & 1.42 & 56.28 & 69.73 & 67.74  & 70.05  \\
\multicolumn{4}{c|}{TDCT} & 1.08 & 61.35 & 81.00 & 78.95 & 82.04 & 1.50 & 54.89  & 69.59 & 67.63 & 70.01 \\
\multicolumn{4}{c|}{STCNet} & 1.20 & 60.95 & 80.98 & 78.89 & 82.05 & 1.57 & 52.39 & 69.36 & 67.69 & 69.65  \\
\bottomrule
\end{tabular}}
\vspace{-0.4cm}
\label{tab6}
\end{table*}

\subsection{Ablation Study}
\subsubsection{Module Ablation}
To evaluate the contribution of each component in our proposed approach, we conduct a series of ablation experiments by progressively integrating each module. The results are shown in Table~\ref{tab5}. We first conduct ablation experiments using two baselines: training without data augmentation and using a simple label-conditioned generation. The results show that data augmentation effectively alleviates the limitations of the training set and improves the performance of gesture recognition. Subsequently, we introduce semantic representations as conditions and adopt the GMSS strategy, which yields significant improvements across all backbones and datasets, demonstrating that the semantic guidance and sampling enhance the faithfulness and diversity of the generated samples. Finally, we incorporate the SASS strategy to explicitly expand the distribution of the training data, which further improves the downstream classification performance. Compared to the baseline, our method achieves an average improvement of \textbf{5.8}\%, confirming its effectiveness and generalizability.
To further analyze model stability for individual subjects under the intra-subject setting, Fig.~\ref{sub_ablation} illustrates the classification accuracy for each subject from the DB7 and DB4 datasets. Our method yields consistent improvements for the majority of subjects, demonstrating its effectiveness in enhancing general performance.

\subsubsection{Hyperparameters Ablation}
To better understand the influence of key hyperparameters on the performance, we conduct ablation studies focusing on four key hyperparameters: the number of iterations (\textit{iter}), the neighborhood radius ($\epsilon$), the neighbor counts $k$ and the confidence threshold for filtering generated samples. The first three parameters jointly control the sparsity level of generated samples: increasing \textit{iter} corresponds to deeper optimization toward sparse regions, while enlarging $\epsilon$ and $k$ expands the neighborhood considered during sampling and thereby produces samples with greater sparsity (e.g., higher FID and lower CAS). The confidence threshold serves as a filter mechanism to avoid introducing low-quality samples. The experimental ablations in Table~\ref{tab6} present the results across different settings on DB7 and DB4. We observe that increasing \textit{iter}, $\epsilon$, and $k$ slightly affects classification accuracy, indicating that downstream performance remains relatively stable across a wide range of sparsity settings. This demonstrates the robustness of SASG-DA to sparsity-controlled generation. Additionally, performance remains consistent across a range of confidence thresholds, indicating that the filtering mechanism does not overly restrict sample diversity. Overall, these results highlight the robustness of our method concerning key hyperparameter choices and confirm that it generalizes well without requiring sensitive tuning.

Based on these observations, the default settings can serve as a robust starting point across different datasets, although the exact optimal values may vary slightly. In practice, the neighborhood radius $\epsilon$ is guided by the latent space scale (e.g., comparable to the average inter-sample distance in the training set), while $k$ and \textit{iter} are related to dataset size and optimization depth (e.g., moderately larger values facilitate sufficient exploration of sparse regions and stable optimization when introducing more samples). Within reasonable ranges, different combinations of these parameters yield comparable downstream performance.

\subsubsection{Semantic Encoder Ablation}

We further assess the impact of task-aware classifiers on the SRG mechanism using different backbones (Crossformer, TDCT and STCNet). As shown in Table~\ref{tab6}, despite differences in test-set performance, the generative metrics show slight variations, while downstream classification performance remains robust, indicating that SRG is not tightly coupled to a specific backbone architecture or to the absolute strength of the pretrained classifier. This behavior can be attributed to the fact that the semantic representations are modeled from the training data, with all three encoders achieving near-saturated training performance and consistently capturing the class-conditional structure. The diffusion model then effectively leverages these representations to generate relatively stable outputs. Crossformer was used as one representative backbone for the main experiments, serving as an alternative among the tested encoders.

\begin{table}[!b]
\vspace{-0.6cm}
\centering
\caption{Quantitative evaluations of generation quality on DB7 and DB4.}
\label{tab7}
\begin{tabular}{ccccccc}
\toprule
  \multirow{2}{*}{Guidance}  &\multicolumn{2}{c}{DB7}  & \multicolumn{2}{c}{DB4}  \\
\cmidrule(lr){2-3} \cmidrule(lr){4-5}
\multicolumn{1}{c}{}  & FID & CAS   & FID & CAS  \\
\midrule
Label Condition  & 2.77 & 49.83  & 2.95 & 45.99  \\
GMSS  & 0.88 & 71.73 & 0.87 & 68.09\\
\bottomrule
\end{tabular}
\end{table}

\section{Further Analysis and Discussion}
\subsection{The Faithfulness of Generated Samples}
To further validate the faithfulness of generated samples, we conduct quantitative evaluations of generation metrics, distributional analyses, and qualitative sample visualizations. As shown in Table~\ref{tab7}, both the Fréchet Inception Distance (FID) and Category Accuracy Score (CAS) consistently demonstrate the effectiveness of the proposed Semantic Representation Guidance mechanism in guiding the generation process. Specifically, the lower FID values indicate that the generated samples are closer to the real data distribution, while higher CAS values reflect improved alignment with class-specific characteristics. These results suggest that incorporating fine-grained semantic representations as conditional information allows the model to synthesize more realistic and discriminative samples for each class.

\begin{figure}[!t]
    \centering
    \label{fig5}
    \begin{tabular}{ccc}
    \hspace{-0.25cm}
    \begin{subfigure}[]{
      \includegraphics[width=0.45\columnwidth]{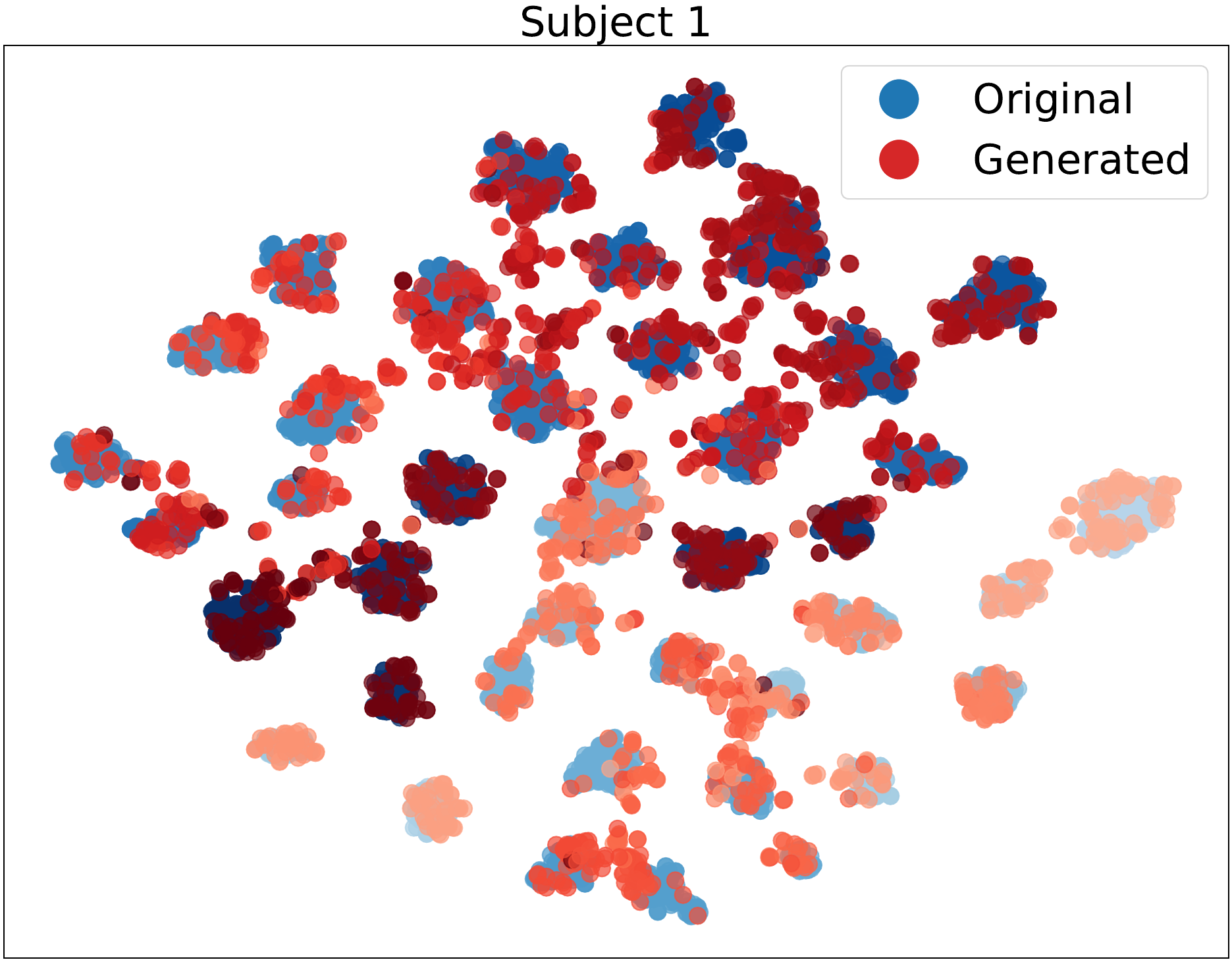}}
    \end{subfigure} &
    \begin{subfigure}[]{
      \includegraphics[width=0.45\columnwidth]{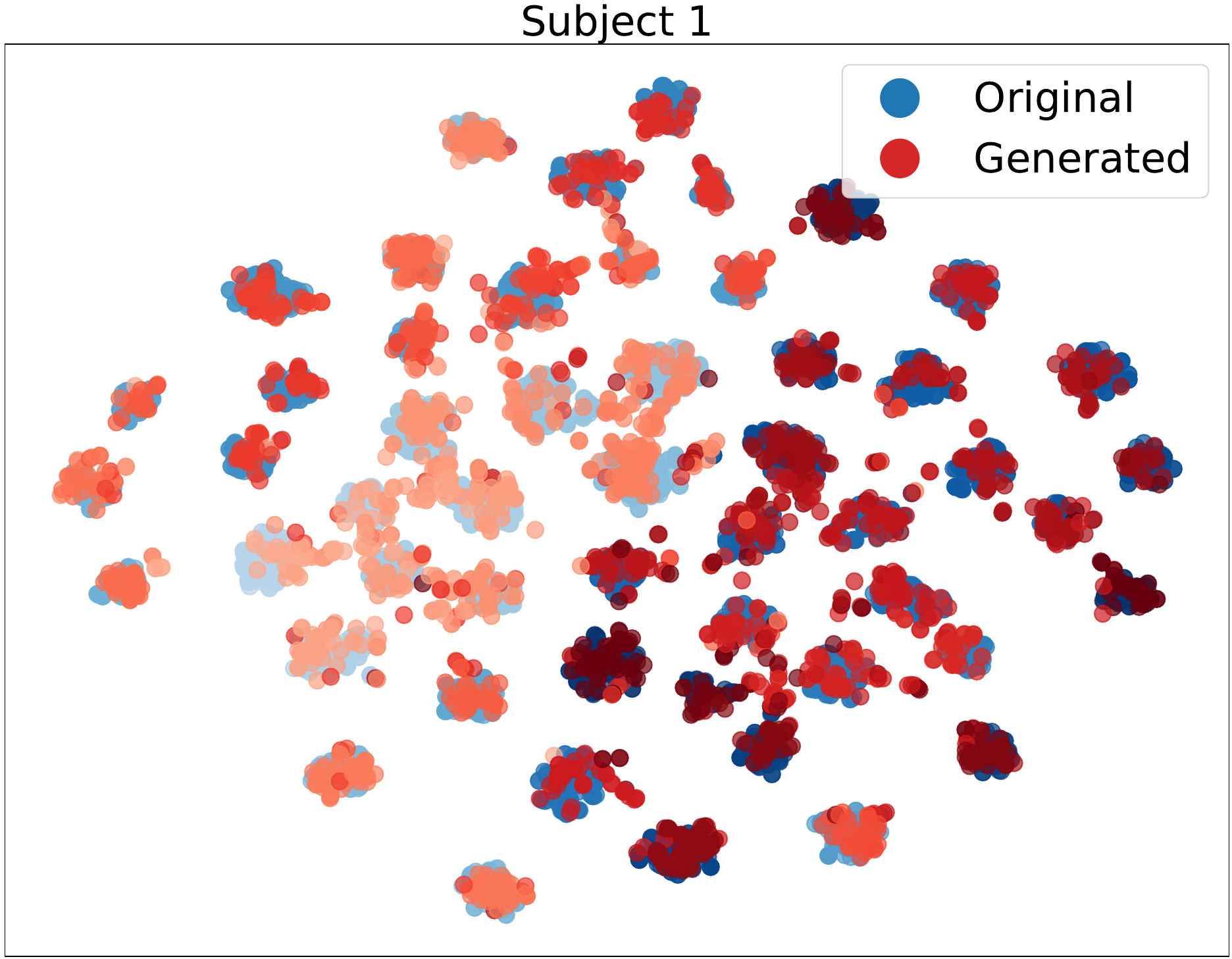}}
    \end{subfigure} &
    \vspace{-0.2cm}
  \end{tabular}
    \caption{Distribution visualizations of the original (blue) and generated (red) samples for subject 1 from (a) DB7 and (b) DB4. Different shades of each color denote different gesture classes.}
    \vspace{-0.5cm}
    \label{distribution}
\end{figure}

To assess distributional faithfulness, we randomly select one subject from DB7 and DB4. Features are extracted from both real and generated samples using a pretrained classifier, and subsequently projected into a two-dimensional space via t-SNE. As shown in Fig.~\ref{distribution}, generated samples closely align with real ones in the embedding space, indicating high distributional similarity. This demonstrates that the proposed Semantic Representation Guidance mechanism effectively preserves the intrinsic data structure and generates class-consistent samples within the real data manifold.

\begin{figure}[!t]
    \centering
    \subfigure[]{
        \includegraphics[width=0.45\columnwidth]{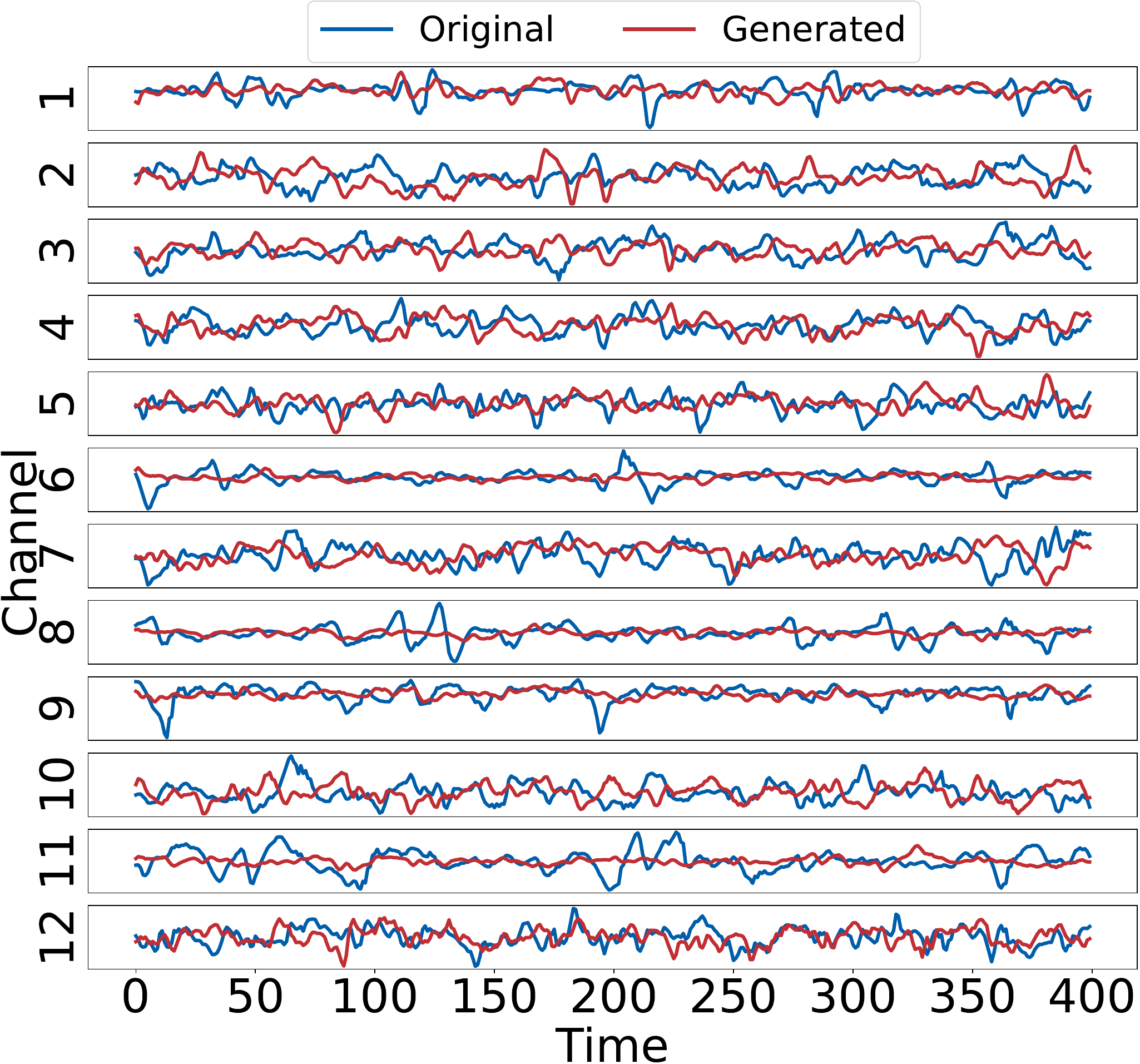}
    }
    \subfigure[]{
        \includegraphics[width=0.45\columnwidth]{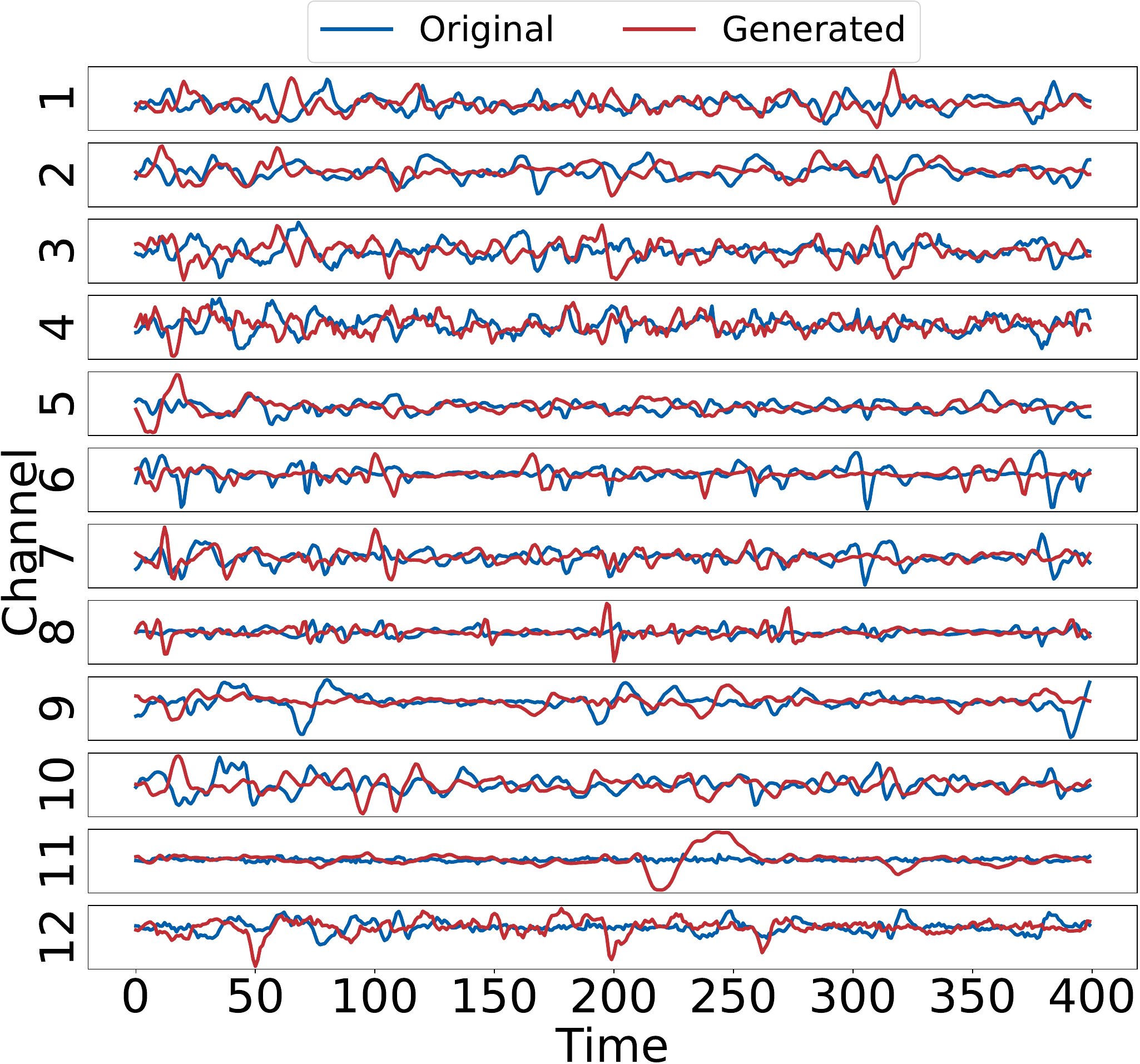}
    }
     \vspace{-0.3cm}
    \caption{Examples of original (blue) and generated (red) samples of (a) DB7 and (b) DB4.}
    \label{samples}
     \vspace{-0.6cm}
\end{figure}

We further provide qualitative comparisons by randomly selecting one sample from each dataset. As illustrated in Fig.~\ref{samples}, the generated sEMG signals exhibit waveform patterns and temporal dynamics that closely resemble those of the corresponding real samples. In particular, even fine-grained signal details, such as subtle fluctuations and transient peaks, are well generated, indicating that the model is capable of generating high-fidelity sEMG signals with refined structure. Further analyses of the physiological plausibility of the synthetic data are presented in the Supplementary Material.

\begin{figure*}[!t]
    \centering
    \subfigure[]{
        \includegraphics[width=0.2\textwidth]{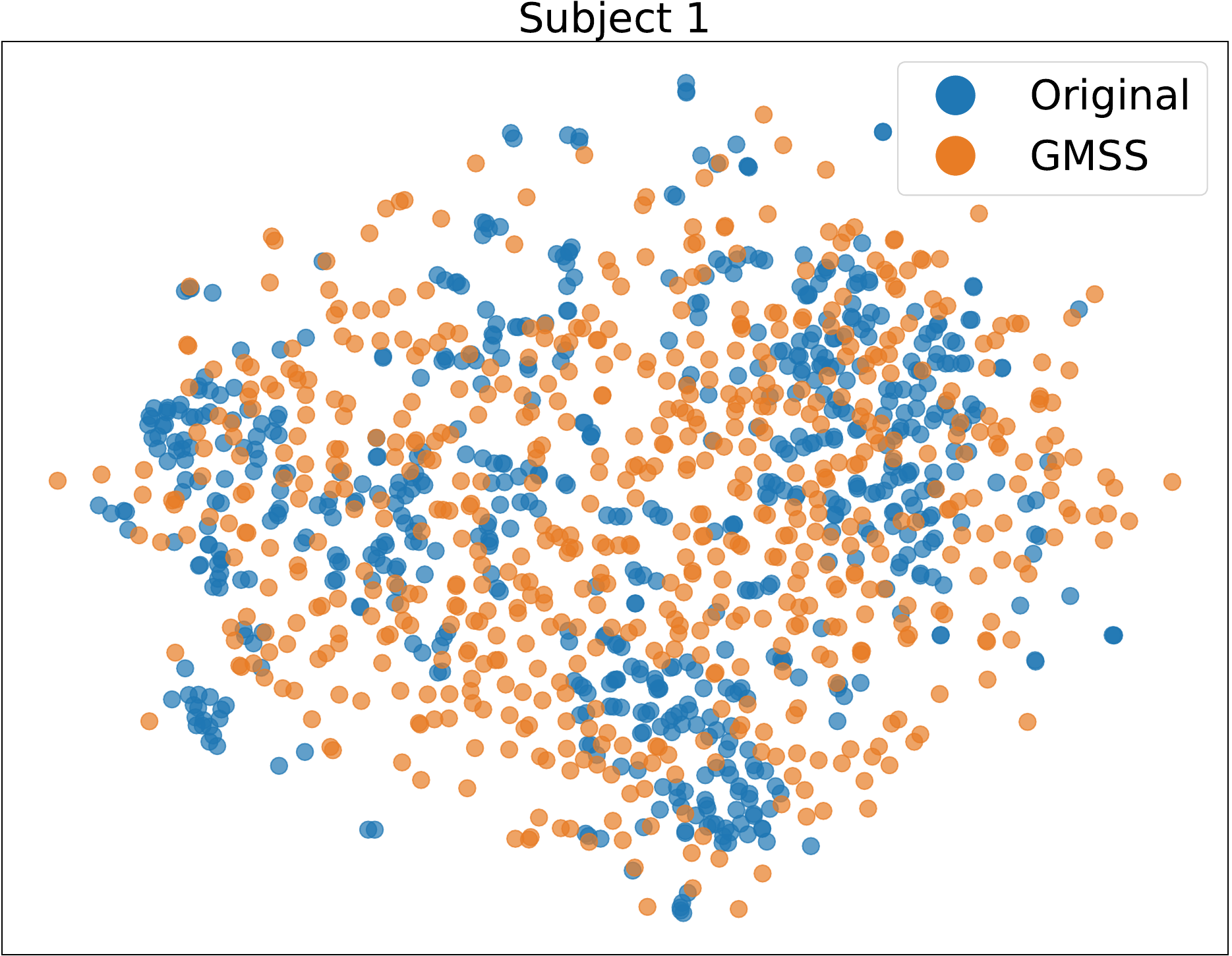}
    }
   \subfigure[]{
        \centering
        \includegraphics[width=0.2\textwidth]{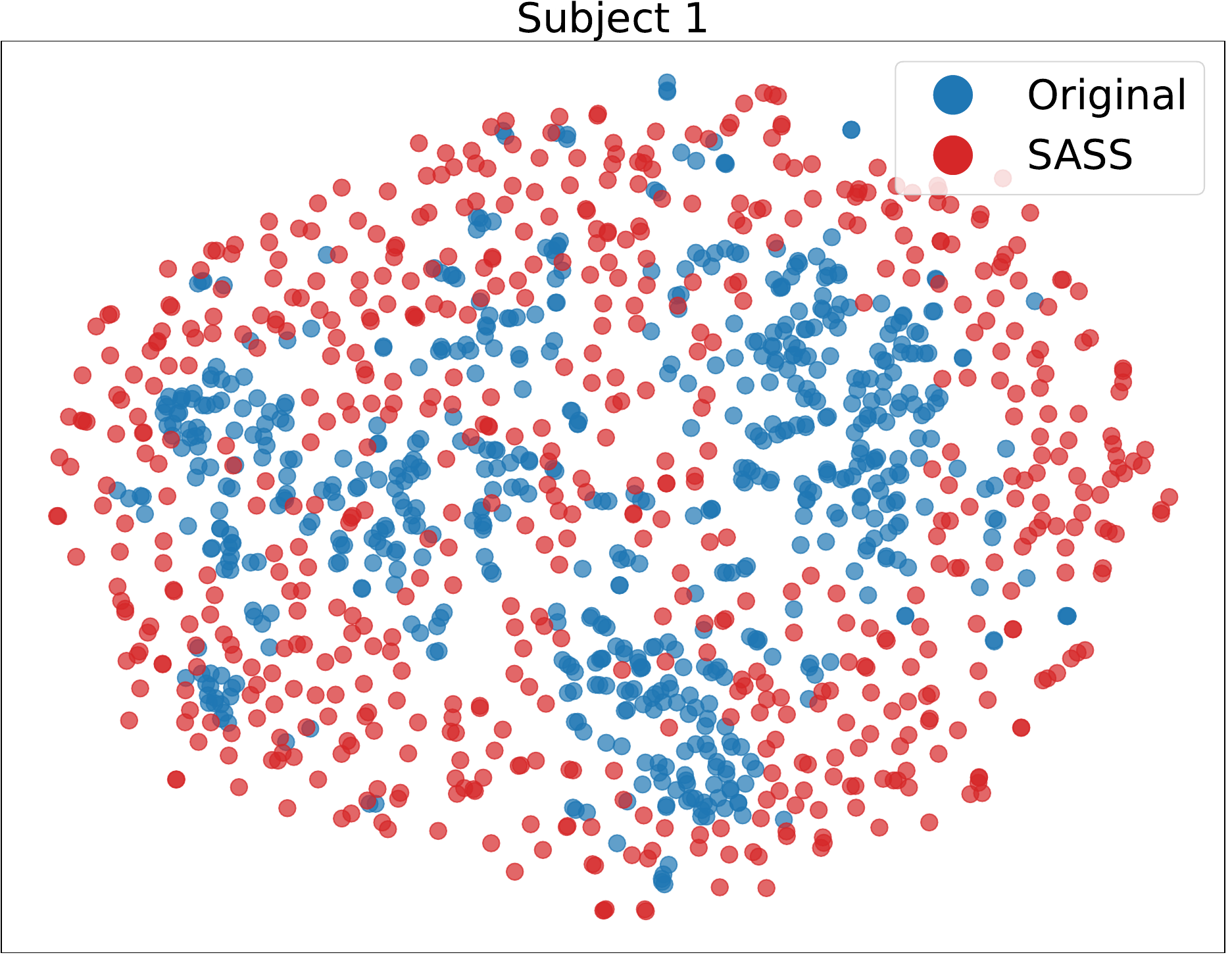}
    }
    \subfigure[]{
        \includegraphics[width=0.2\textwidth]{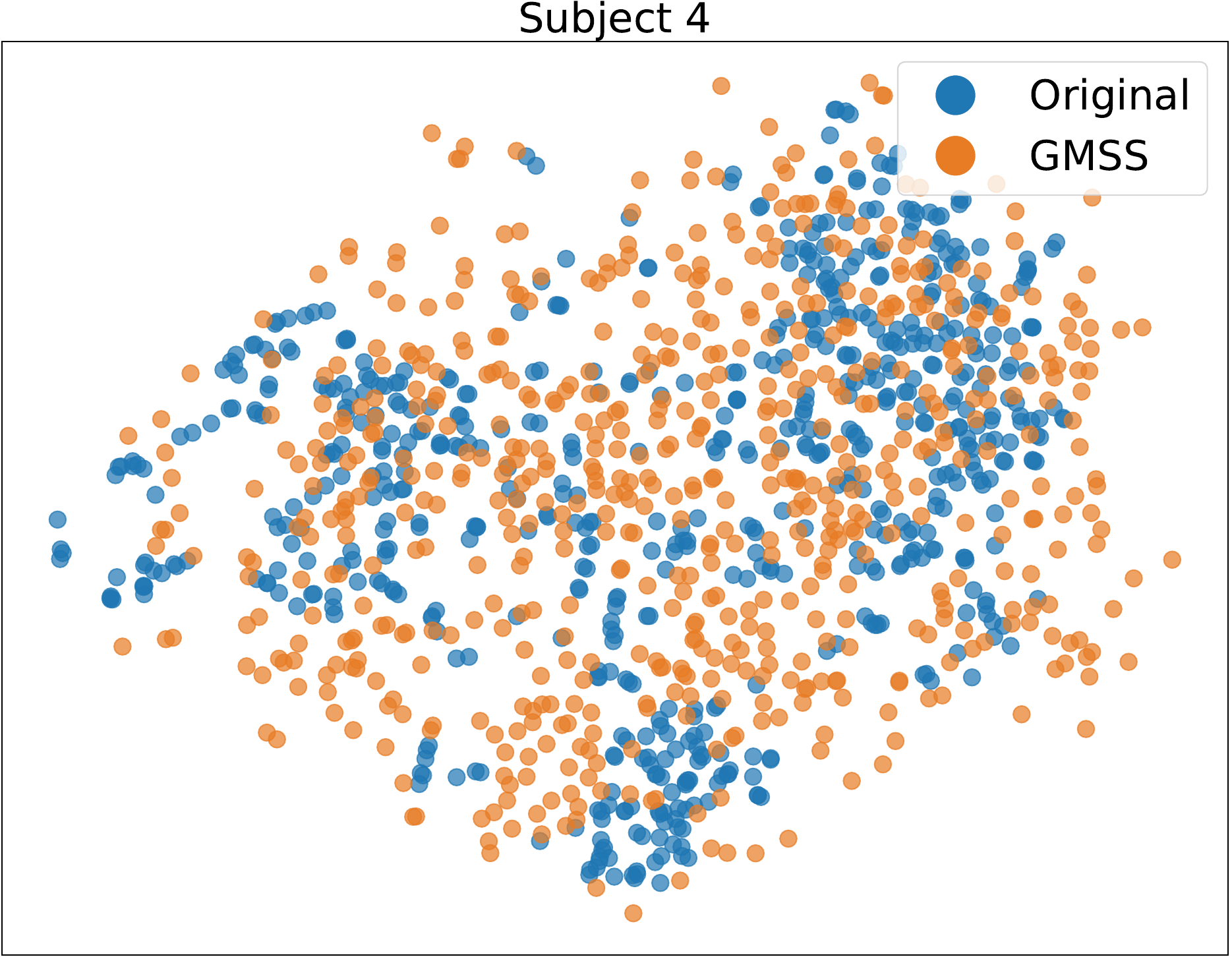}
    }
   \subfigure[]{
        \centering
        \includegraphics[width=0.2\textwidth]{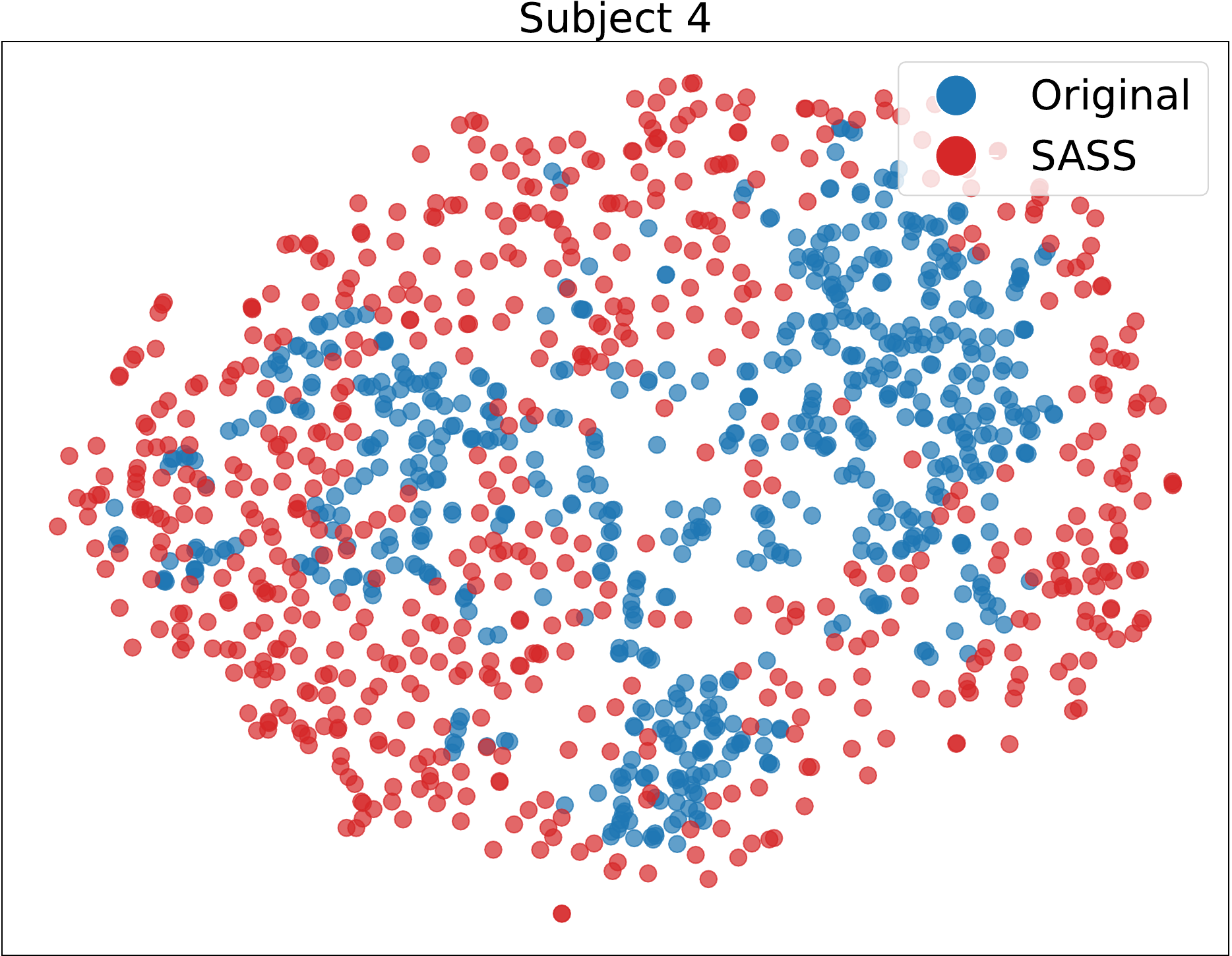}
    }
    \subfigure[]{
        \includegraphics[width=0.2\textwidth]{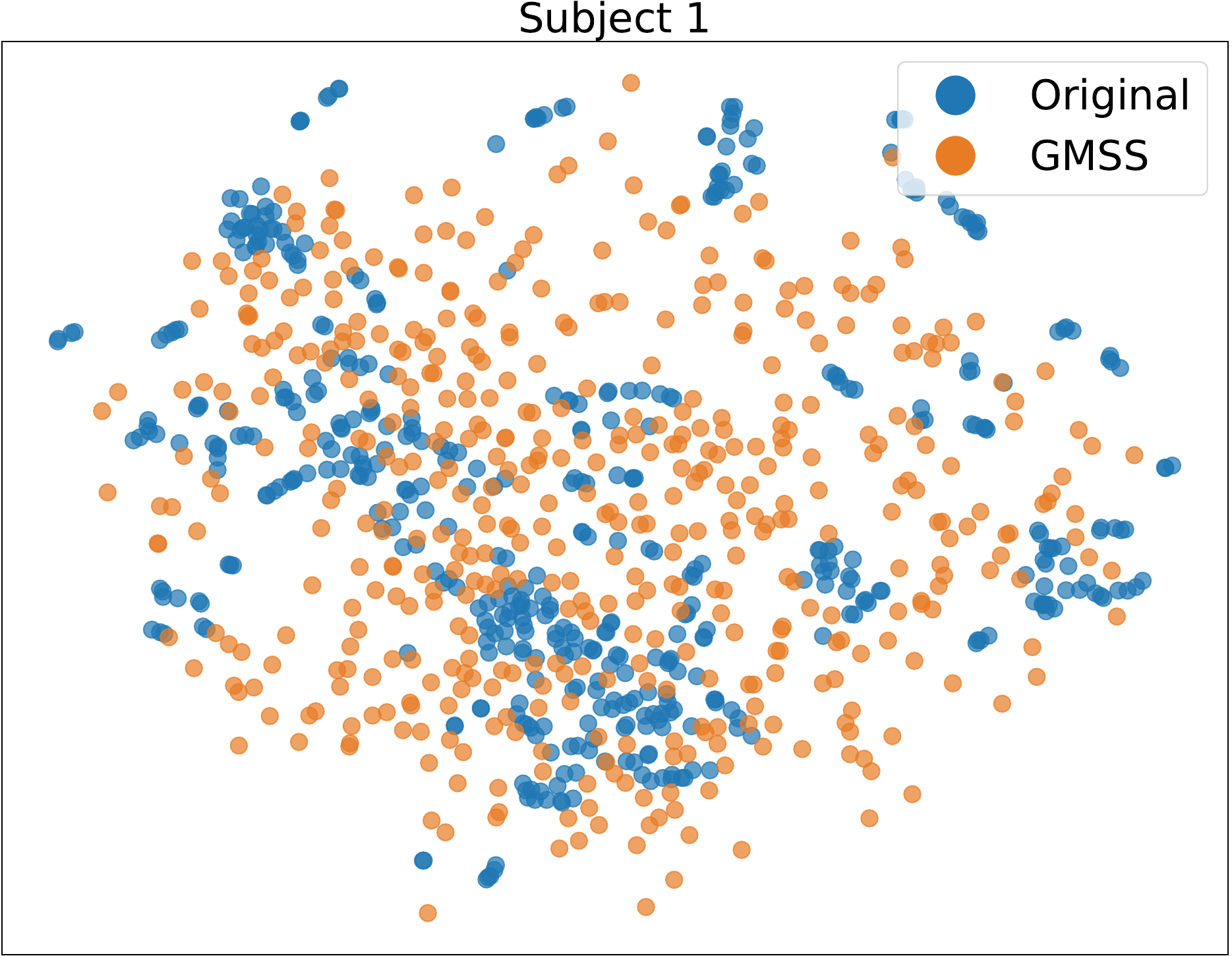}
    }
   \subfigure[]{
        \centering
        \includegraphics[width=0.2\textwidth]{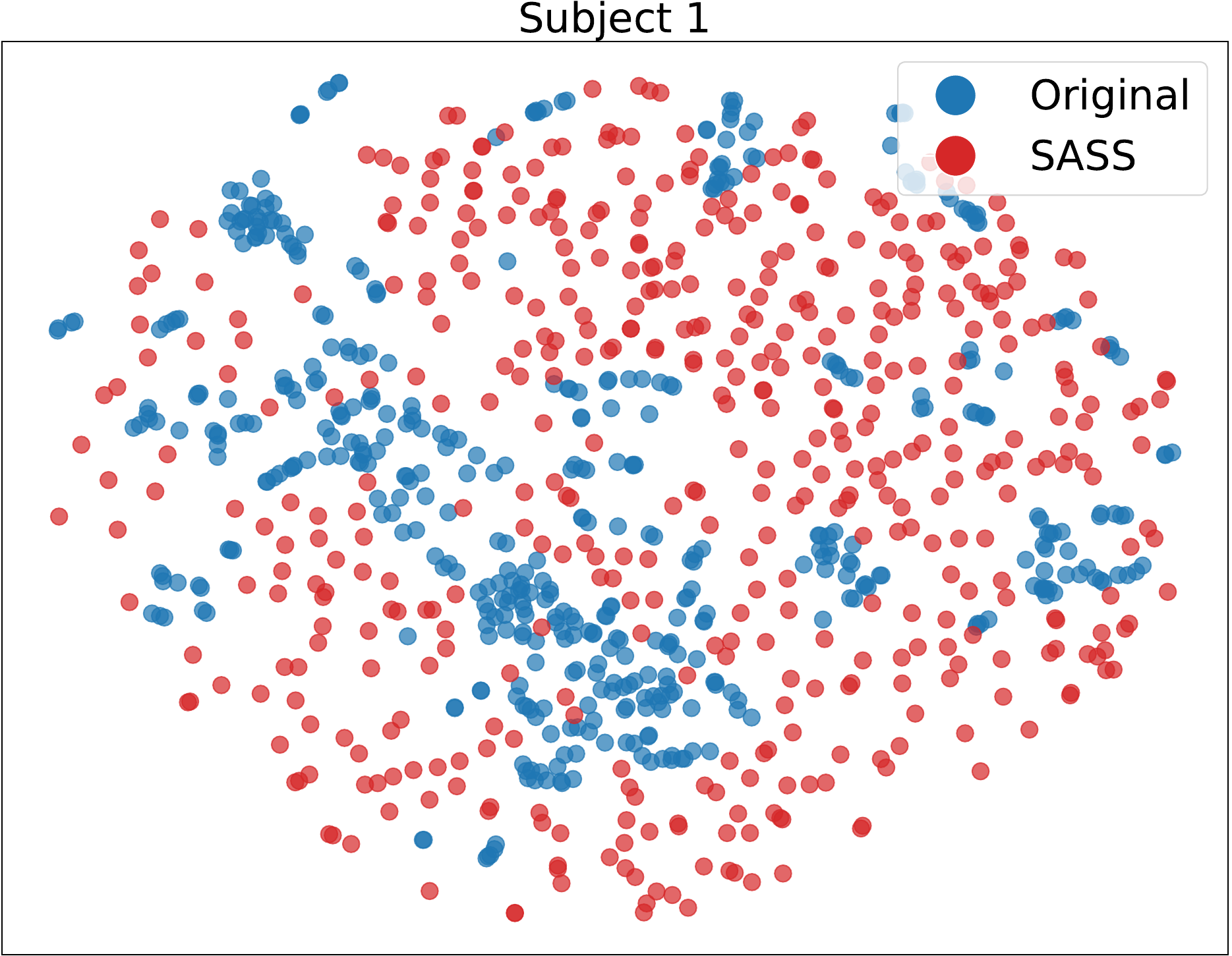}
    }
    \subfigure[]{
        \includegraphics[width=0.2\textwidth]{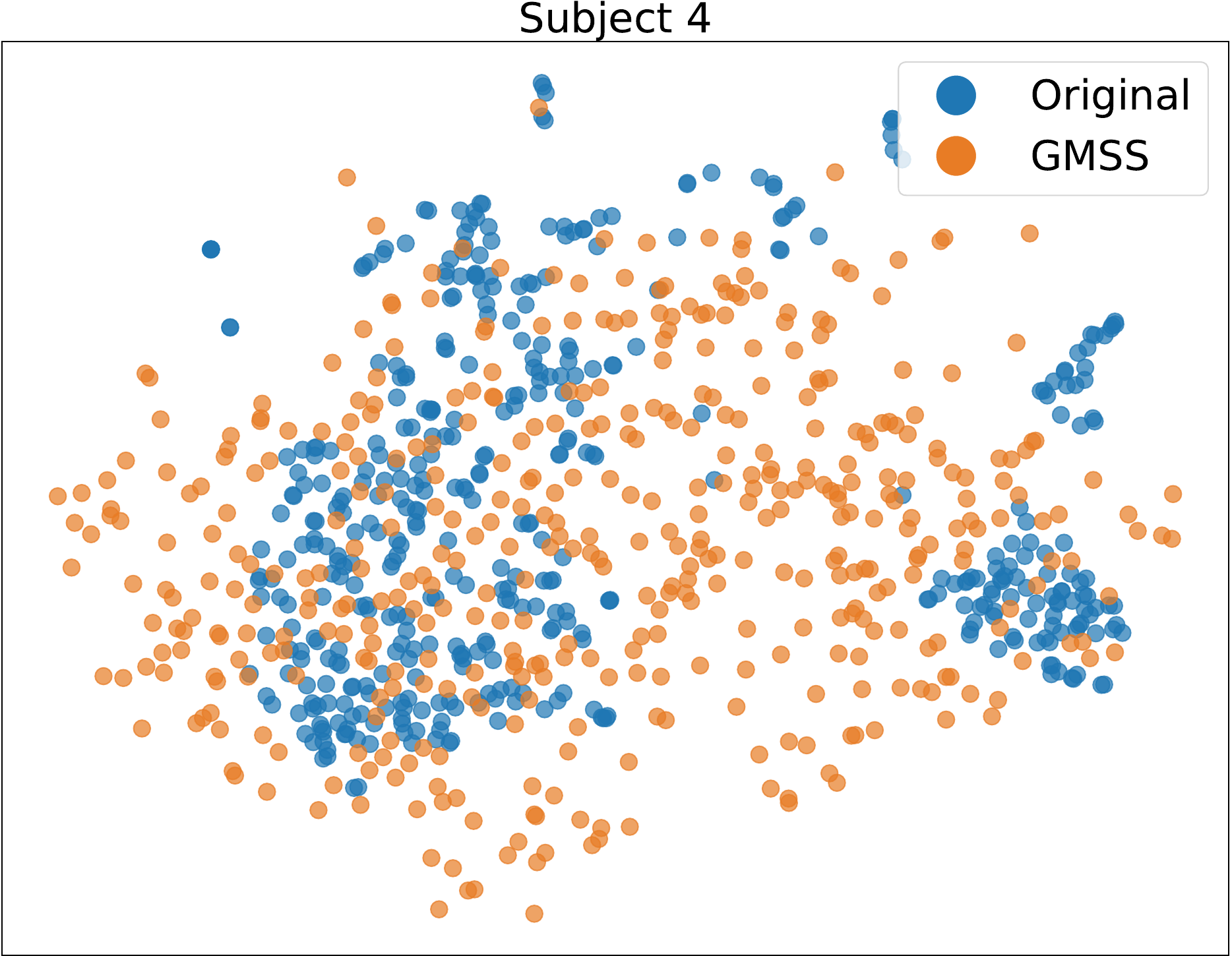}
    }
   \subfigure[]{
        \centering
        \includegraphics[width=0.2\textwidth]{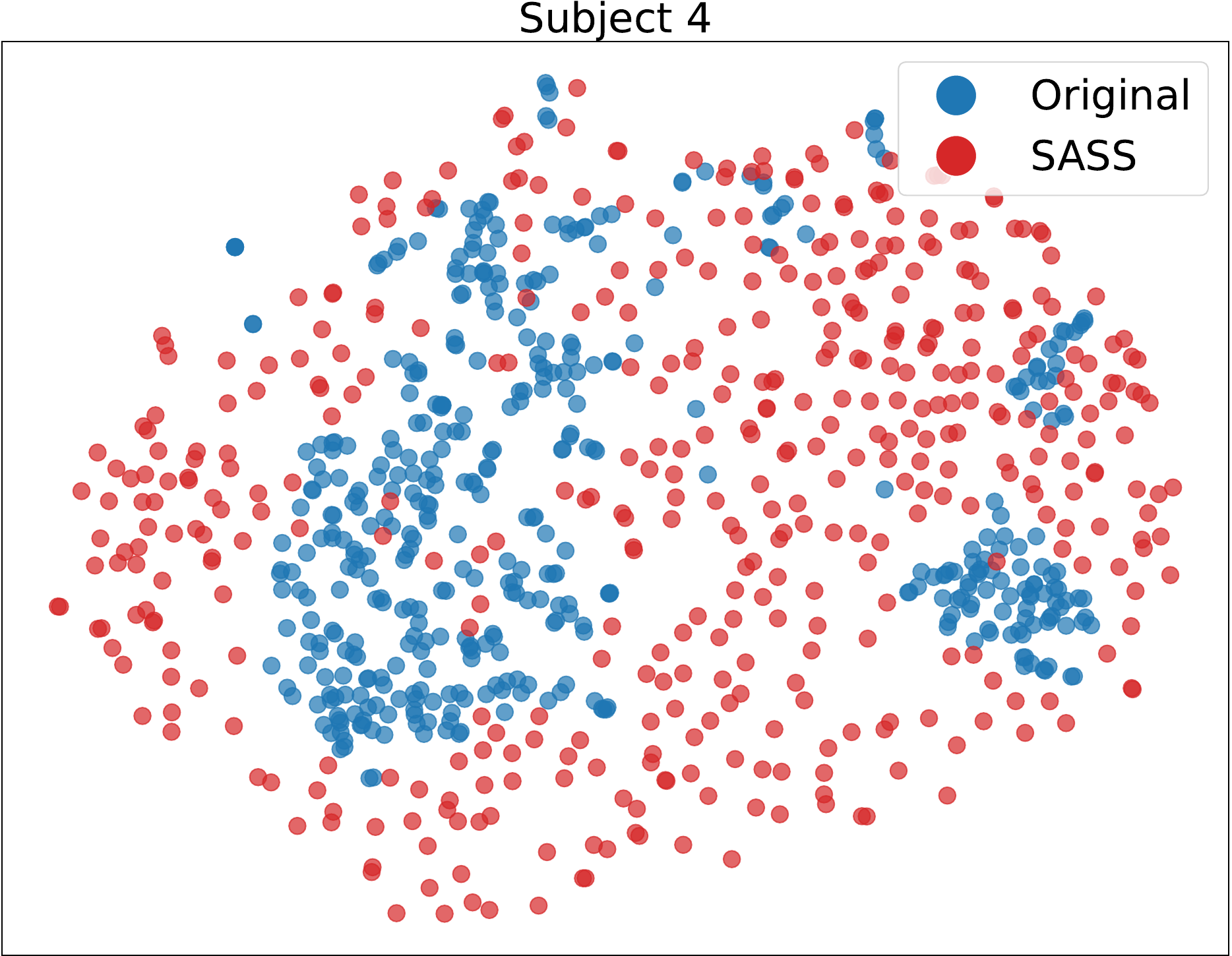}
    }
     \vspace{-0.2cm}
    \caption{Condition visualization to show the effectiveness of the SASS mechanism for subjects 1 and 4 from (a-d) DB7 and (e-h) DB4.}
     \vspace{-0.4cm}
    \label{fig6}
\end{figure*}

\begin{figure*}[!t]
    \centering
    \subfigure[]{
        \includegraphics[width=0.25\textwidth]{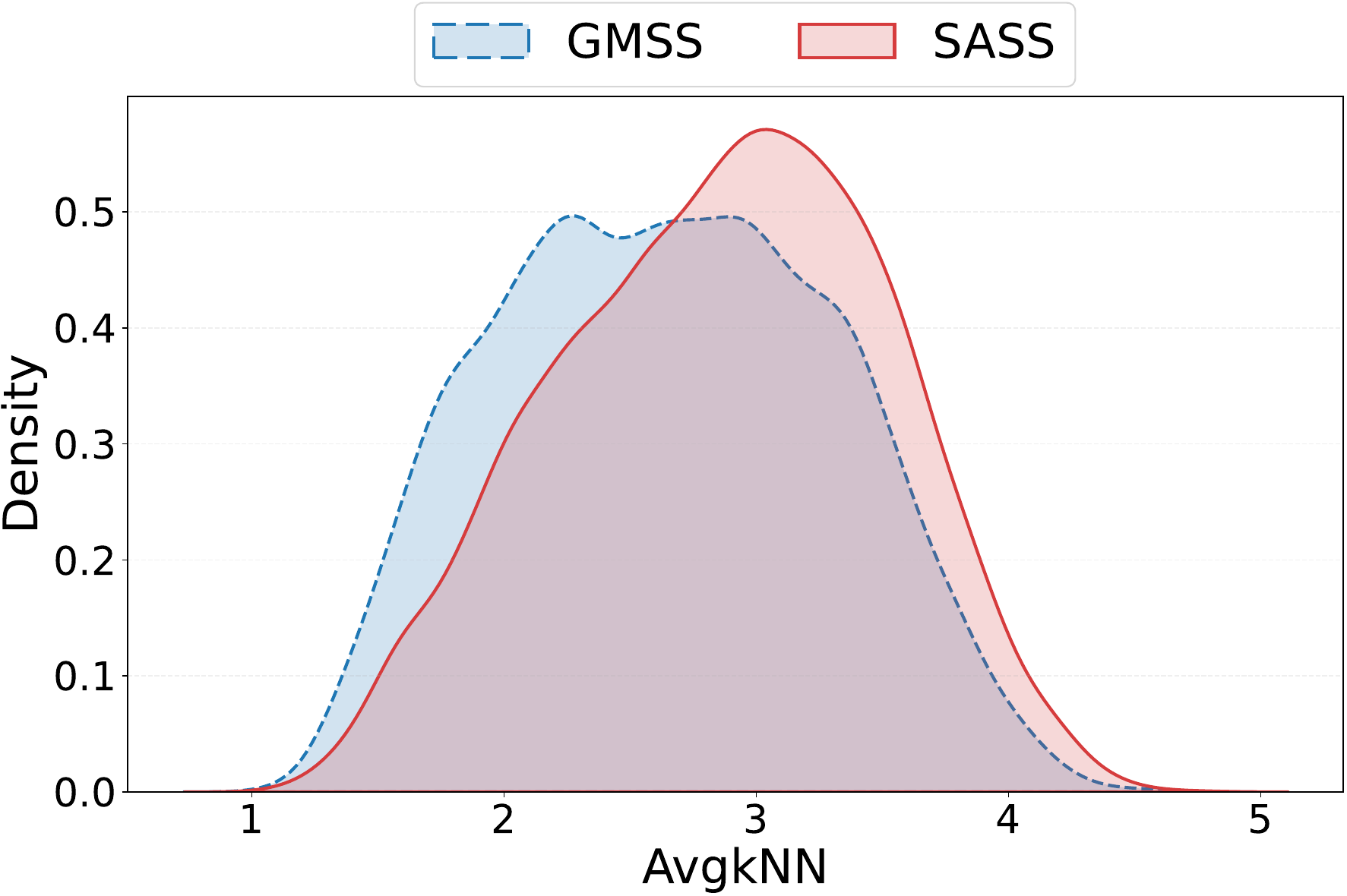}
    }
    \subfigure[]{
        \includegraphics[width=0.25\textwidth]{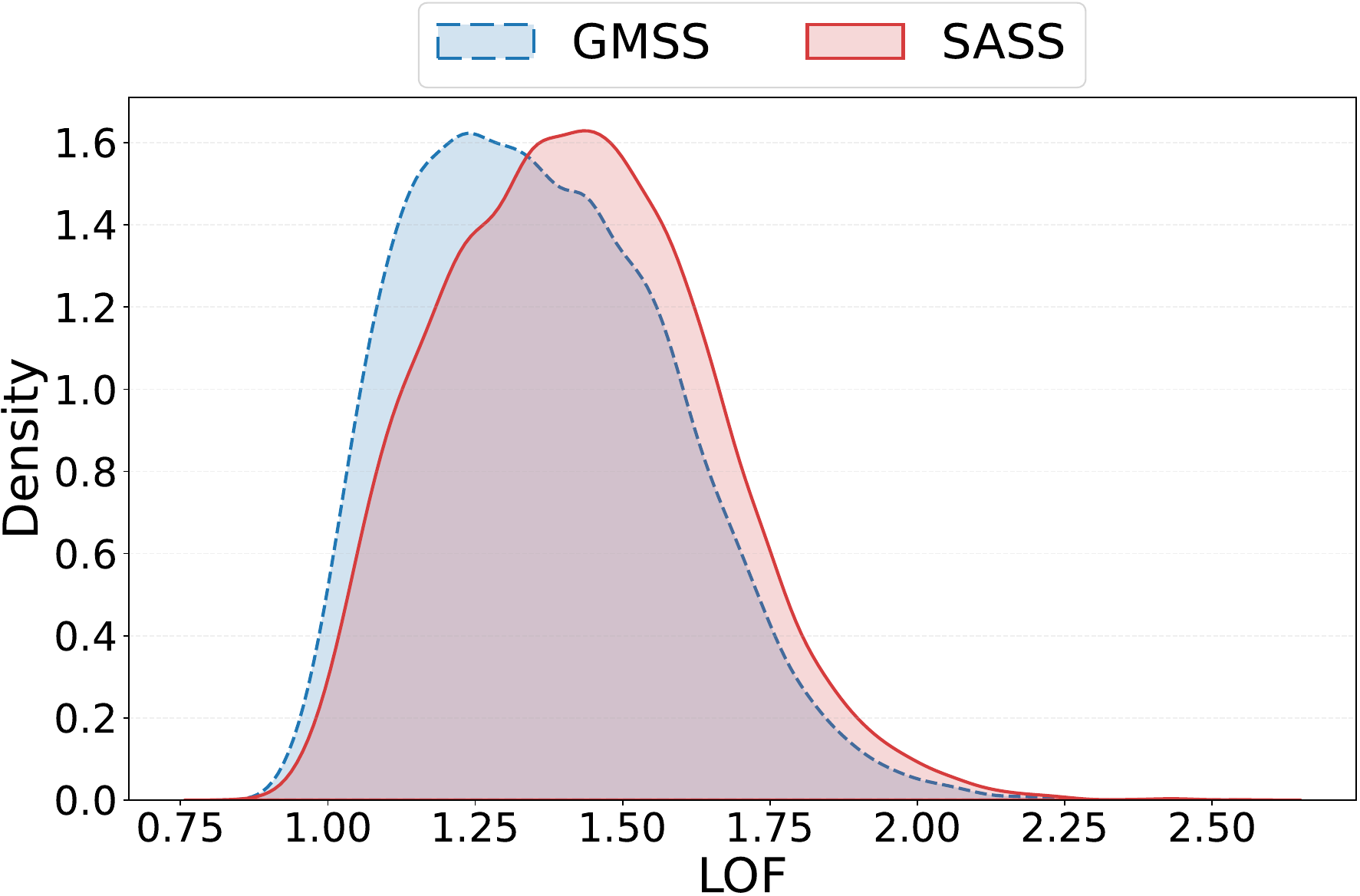}
    }
    \subfigure[]{
        \includegraphics[width=0.25\textwidth]{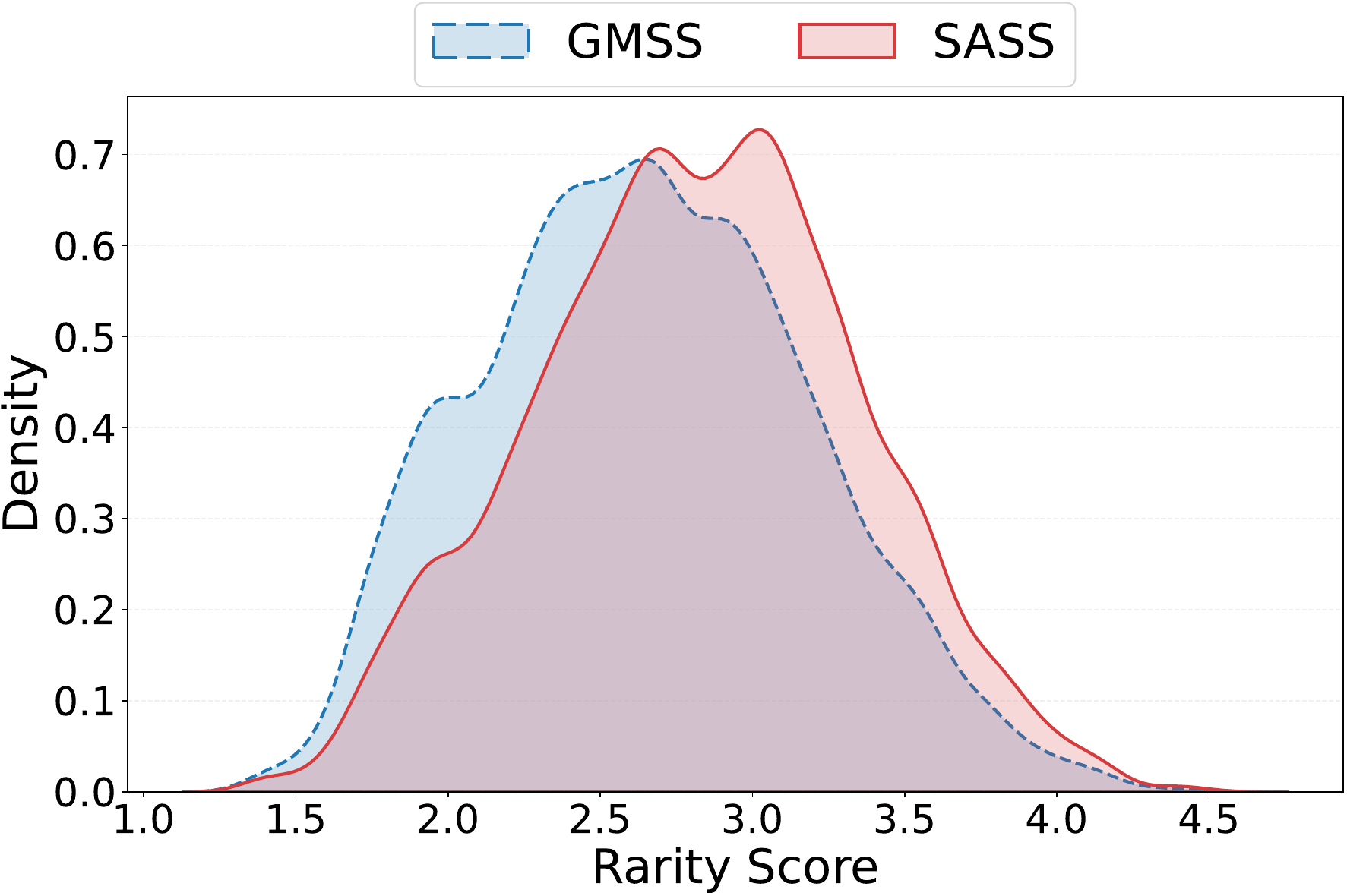}
    }
     \vspace{-0.2cm}
    \caption{Quantitative evaluation of the generated sample sparsity on DB7 using three different metrics: (a) AvgKNN, (b) LOF, and (c) Rarity score.}
    \label{sparsity}
    \vspace{-0.6cm}
\end{figure*}

\subsection{Extension of Dataset Coverage and Downstream Utility}

To verify that our SASS strategy explicitly expands the data distribution, we conduct condition space visualization, quantitative sparsity evaluations and downstream utility analysis. Specifically, we use t-SNE to project the semantic representations of the original training set, as well as those obtained through the GMSS and SASS mechanisms, into a two-dimensional space. As shown in Fig.~\ref{fig6}, the semantic representations generated by the GMSS mechanism are evenly distributed within the original feature space, producing diverse and distribution-consistent conditions. Furthermore, the semantic representations generated by the SASS mechanism are distributed in the sparser regions of the original feature space, which significantly enlarges the coverage of the semantic space while maintaining consistency with the original data manifold.

We further verify that the samples generated under sparse condition guidance are indeed sparse in nature through quantitative metrics. Specifically, following the settings in~\cite{sehwag2022generating, um2024self}, we adopt three sparsity measures on generated samples: the average k-nearest neighbor distance (AvgKNN), the Local Outlier Factor (LOF), and the \textit{rarity score}. As shown in Fig.~\ref{sparsity}, the generated samples guided by the sparse conditions consistently exhibit higher AvgKNN distances, larger LOF values, and higher \textit{rarity scores} compared to those generated by the GMSS. These results quantitatively demonstrate that the SASS mechanism successfully shifts the generated samples towards the sparse regions of the data distribution, explicitly expanding the coverage of the training manifold. This property is particularly beneficial for enhancing the generalization capability of the classifier by exposing it to more diverse and underrepresented regions of the original data space.

\begin{figure}[!b]
    \vspace{-0.7cm}
    \centering
    \subfigure[]{
    \includegraphics[width=0.45\columnwidth]{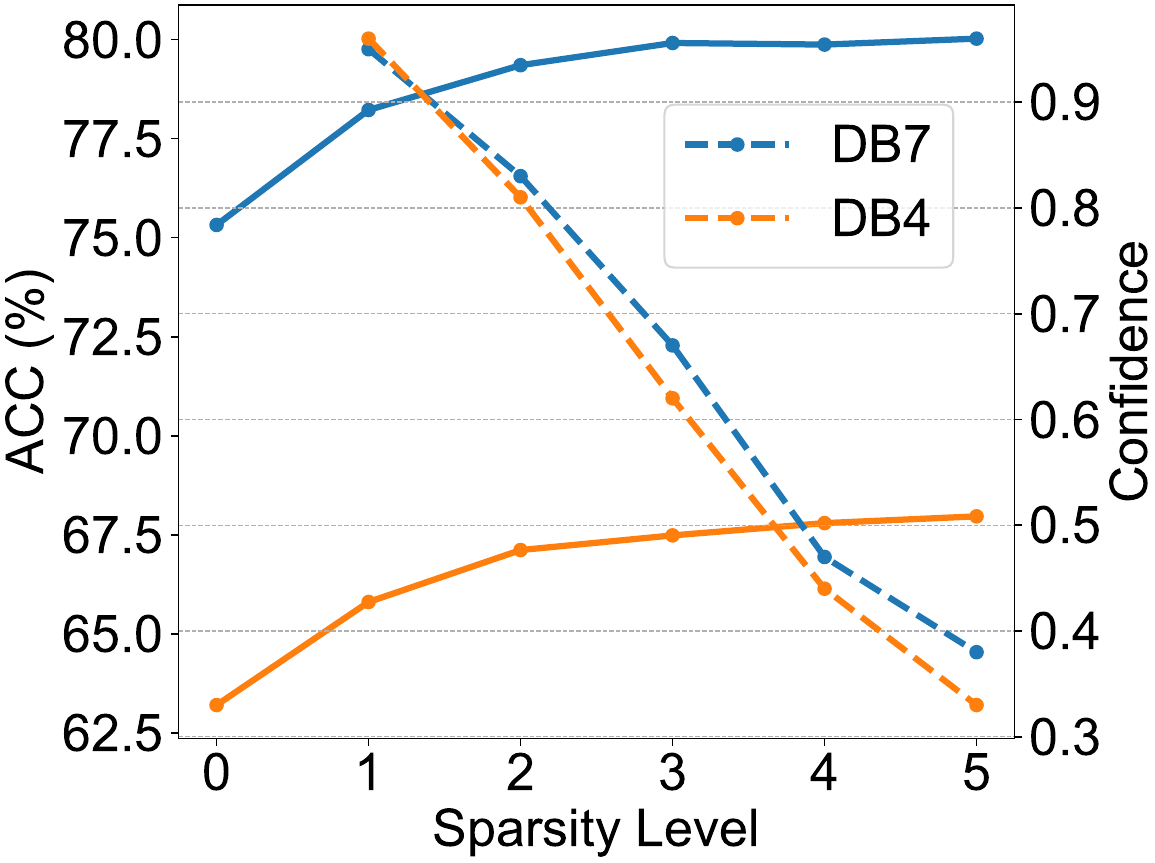}}
    \subfigure[]{
    \includegraphics[width=0.45\columnwidth]{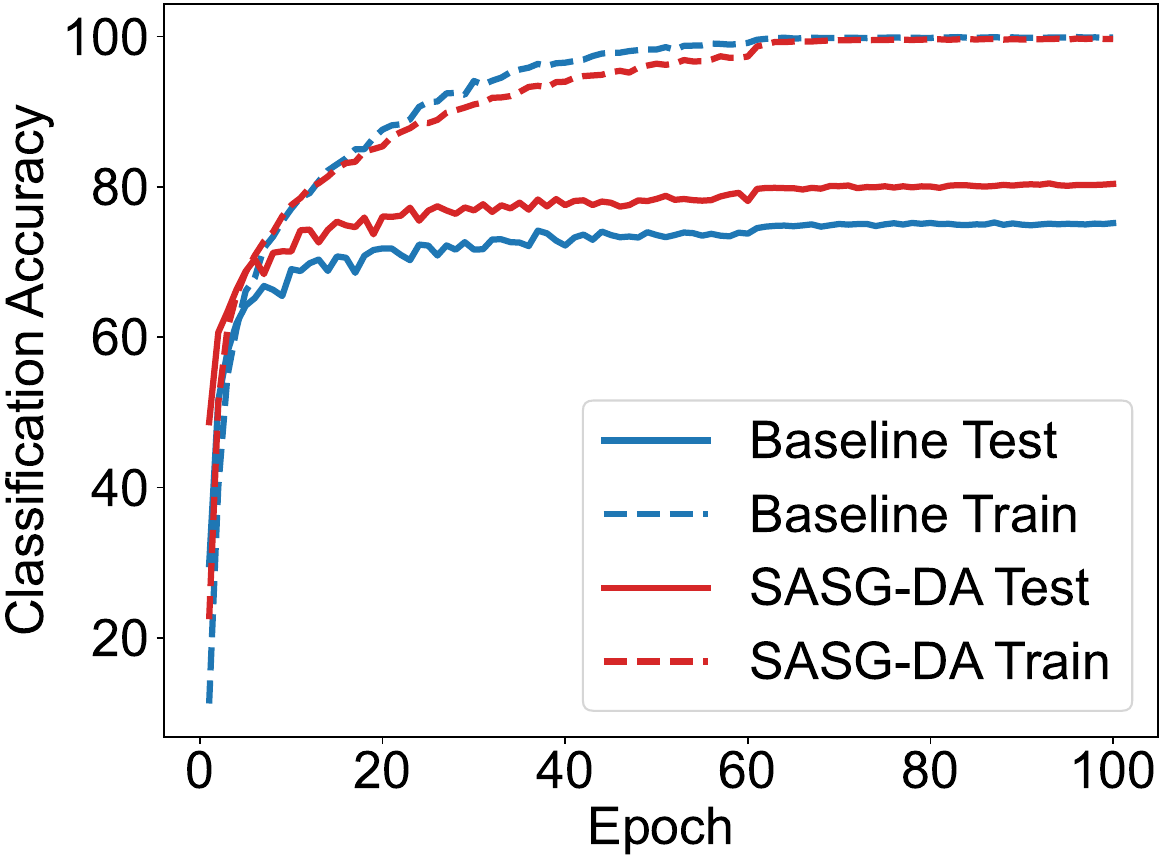}}
     \vspace{-0.2cm}
    \caption{Downstream utility validation by (a) Performance trend across sparsity levels and confidence for generated samples. (b) Training dynamics for a randomly selected subject from DB7.}
    \label{sparsity_level}
\end{figure}

Beyond overall classification performance, we further validate the downstream utility in generating targeted diverse samples by performing sparsity-stratified training analysis. Specifically, we mix samples generated by GMSS and SASS to form a broad distributional range, and then divide them into five equally sized levels based on their \textit{rarity scores}. Level 0 represents the baseline without augmentation. We evaluate each sparsity level on DB7 and DB4. The results in Fig.~\ref{sparsity_level} show that downstream performance consistently improves as sparsity increases and stabilizes at a high level, indicating that the SASS mechanism effectively enhances the utility of generated samples for downstream tasks by enhancing their targeted diversity. While the performance gain gradually saturates at Level 5, highly sparse samples remain valid and do not cause noticeable performance degradation. This observation is consistent with our motivation of explicitly expanding the training distribution to improve downstream utility.

We additionally analyze generated sample confidence across sparsity levels, showing that confidence decreases as sparsity increases, consistent with samples from underrepresented regions being less confidently predicted by the pretrained classifier on the original training set. Furthermore, the training curves in Fig.~\ref{sparsity_level} show that our approach narrows the gap between training and test accuracy compared to the baseline, suggesting mitigated overfitting and improved generalization.

\subsection{Trade-off Between Faithfulness and Diversity}

To illustrate the trade-off between faithfulness and diversity, we compare our method with three common conditional generation baselines: Label Condition, Classifier Guidance~\cite{dhariwal2021diffusion}, and Classifier-Free Guidance~\cite{ho2021classifier}. The results are shown in Table~\ref{tab8}. The Label Condition approach introduces only coarse-grained class information, which results in relatively high FID scores (e.g., 2.77 on DB7) and low CAS scores (49.83), indicating insufficient faithfulness to the target classes. In contrast, Classifier Guidance employs an explicit classifier to steer the generation toward class-consistent samples~\cite{dhariwal2021diffusion}, significantly improving CAS (e.g., up to 86.08) and reducing FID. However, stronger classifier gradient guidance can overly constrain the generation process, limiting the diversity of the synthesized samples by aligning them too closely with the classifier’s decision boundaries. Classifier-Free Guidance avoids the need for an external classifier by enabling the generative model to produce both conditional and unconditional outputs during inference~\cite{ho2021classifier}. While it avoids the overhead of an external classifier, stronger condition strength still tends to reduce diversity by pulling generated samples toward the high-density regions of the class distribution.

\begin{table*}[thbp]
\centering
\caption{Performance comparison of different guidance strategies on DB7 and DB4. FID and cFID values are reported with cFID computed conditionally per class. For clarity, cFID A-T refers to the Augmented training set vs the Test set.}
\resizebox{\textwidth}{!}{%
\begin{tabular}{c|cccccc|cccccc}
\toprule
\multirow{2}{*}{Method} 
& \multicolumn{6}{c}{DB7 (cFID Train-Test=5.97)} 
& \multicolumn{6}{|c}{DB4 (cFID Train-Test=8.64)}  \\
\cmidrule{2-7} \cmidrule{8-13}
  & FID & CAS & cFID A-T & Crossformer & TDCT & STCNet 
& FID & CAS & cFID A-T & Crossformer & TDCT & STCNet\\
\midrule
Label Condition & 2.77 & 49.83 & 5.15 & 80.49 & 77.09 & 80.62 & 2.95 & 45.99 & 8.24 & 68.81 & 66.04 & 68.19    \\
Classifier Guidance & 1.73 & 86.08 & 4.83 & 79.72 & 76.82 & 80.11 & 2.03 & 92.18 & 7.38 & 67.81  & 65.23  & 68.21  \\
Classifier-Free Guidance & 1.62 & 77.00 & 4.97 & 80.54 & 77.46  & 80.95  & 2.13 & 71.44 & 7.15 & 68.91 & 66.30  & 68.43  \\
GMSS & 0.88 & 71.73 & 4.27 & 80.87 & 77.57 & 81.67 & 0.87 & 68.09 & 5.75 & 69.44 & 66.99 & 69.41  \\
SASG-DA & 1.35 & 58.65 & 4.23 & 81.31 & 78.77 & 82.15 & 1.42 & 56.28 & 5.73 & 69.73  & 67.74 & 70.05  \\

\bottomrule
\end{tabular}}
\vspace{-0.4cm}
\label{tab8}
\end{table*}

\begin{table*}[!t]
\centering
\caption{Comparison of computational cost, method characteristics, and performance gain across datasets.}
\renewcommand{\arraystretch}{0.97}
\resizebox{0.85\textwidth}{!}{
\begin{tabular}{lccccccc}
\toprule
Method &
\makecell{FLOPs\\(G)} &
\makecell{Params\\(M)} &
\makecell{Training\\Time (s/it)} &
\makecell{Inference Throughput\\(samples/s)} &
Faithfulness &
Diversity &
\makecell{Perf Gain\\(DB7 / DB4 / DB2)} \\
\midrule
E-TRGAN & 0.51 & 2.28 & 0.02 & 476.3 & $\checkmark$ &  & 0.54 / 0.68 / 0.48 \\
DiffMix & 2.17 & 3.60 & 0.23 & 13.1 &  & $\checkmark$ & 0.57 / 0.60 / 0.10 \\
SSE-LDM & 0.40 & 1.36 & 0.08 & 21.0 & $\checkmark$ &  & 0.45 / 0.85 / 0.19 \\
EEG-Enhancing & 2.17 & 3.60 & 0.20 & 12.9 & $\checkmark$ &  & 0.24 / 0.47 / 0.26 \\
NTD & 0.37 & 0.56 & 0.04 & 25.1 & $\checkmark$ &  & 2.42 / 2.74 / 1.43 \\
PatchEMG & 4.34 & 3.60 & 0.38 & 7.6 & $\checkmark$ &  & 3.23 / 3.33 / 2.20 \\
DESAM & 2.11 & 2.18 & 0.21 & 10.48 & $\checkmark$ &  & 3.04 / 1.98 / 1.67 \\
Minority & 2.17 & 3.60 & 0.23 & 9.8 &  & $\checkmark$ & 3.24 / 3.32 / 2.45 \\
CADS & 2.17 & 3.60 & 0.23 & 12.8 &  & $\checkmark$ & 4.07 / 4.10 / 3.12 \\
\midrule
\textbf{SASG-DA (500)} &2.26 & 3.60 & 0.25 & 12.7 & $\checkmark$ & $\checkmark$ & 5.70 / 5.96 / 4.73 \\
\textbf{SASG-DA (200)} &2.26 & 3.60 & 0.25 & 32.4 & $\checkmark$ & $\checkmark$ & 5.30 / 5.85 / 4.38\\
\textbf{SASG-DA (50)} &2.26 & 3.60 & 0.25 & 129.3 & $\checkmark$ & $\checkmark$ & 5.17 / 5.67 / 4.36 \\
\bottomrule
\end{tabular}}
\vspace{-0.6cm}
\label{tab_cost}
\end{table*}

Interestingly, higher faithfulness (i.e., lower FID and higher CAS) does not always translate to better downstream classification performance. For instance, although Classifier Guidance achieves the highest CAS, its accuracy on downstream tasks does not outperform the Label Condition. Our approach, SASG-DA, achieves a balanced FID and CAS while delivering consistently superior classification accuracy across multiple backbones. This demonstrates that our method effectively balances the trade-off between generating faithful and diverse samples, which is crucial for maximizing the benefit of data augmentation in downstream applications.

To further examine whether sparsely generated samples plausibly extend the training distribution, we analyze distributional coverage with respect to a held-out test set. Specifically, we compute cFID (class-wise FID), which averages per-class distances to better capture class-level alignment between augmented and test samples. We evaluate cFID between (i) the original training and test sets, and (ii) the augmented training set (Train + Synthetic) and the test set. As shown in Table~\ref{tab8}, incorporating sparse samples consistently reduces the cFID gap, indicating that the generated samples expand the data manifold toward a more plausible distribution.

\subsection{Overall Evaluation on Computational Efficiency and Performance}
To provide a comprehensive evaluation of the proposed method, we compare the computational cost, method characteristics, and performance gains of generative augmentation approaches in Table~\ref{tab_cost}. For fairness, all methods are evaluated under the same settings described in Section~\ref{settings}. FLOPs are reported per diffusion step or forward, training time per iteration depends on the batch size, and inference throughput varies with the number of sampling steps. Here, we focus on the relative computational overhead introduced by our method.

Compared with SOTAs built upon standard diffusion models (e.g., DiffMix~\cite{wang2024enhance}, CADS~\cite{sadat2024cads}), our approach introduces only limited additional computational overhead while yielding consistent and notable performance improvements across three datasets. Importantly, it explicitly enhances faithfulness and diversity at minor cost, without relying on the costly CFG~\cite{ho2021classifier} or other gradient-based guidance~\cite{um2024self}. Methods such as SSE-LDM~\cite{aristimunha2023synthetic} and NTD~\cite{vetter2024generating} achieve low computational cost through specially designed lightweight architectures; however, their performance improvements remain relatively modest. Compared with GAN-based approaches, our method benefits from the inherent stability and strong generative capacity of diffusion models, resulting in more substantial gains in downstream tasks. In addition, we evaluate the efficiency–performance trade-off under reduced sampling steps. Notably, fewer sampling steps substantially increase inference throughput while causing only minor performance loss, showing that our method remains effective in more efficient settings and holds the potential for practical applications. In practice, SASG-DA performs offline augmentation with a one-time generation cost of approximately 20, 7, or 2 minutes per subject for 500, 200, or 50 steps, respectively, generating around 15k samples on large-scale Ninapro datasets, without requiring additional generative computation during downstream training. Compared with single-sample augmentation strategies, SASG-DA provides more consistent gains across backbones and datasets, particularly under cross-subject settings, highlighting a practical trade-off where moderate offline cost yields improved generalization through continuous and reusable high-quality synthetic samples.

In addition, SASG-DA consistently improves performance across all three datasets and multiple backbone architectures, with average gains of 5.70\%, 5.96\%, and 4.73\%, respectively. The variations in improvement are expected given differences in signal characteristics, gesture complexity, and baseline performance, yet the results clearly demonstrate the effectiveness of SASG-DA across diverse data characteristics.

\subsection{Cross-subject Evaluation}

To further evaluate the adaptability of the proposed SASG-DA, we conduct a cross-subject evaluation on the GrabMyo dataset~\cite{pradhan2022multi}. As a relatively recent wrist-based dataset, it offers a representative and practical setting for cross-subject evaluation, consistent with current trends in wearable sEMG~\cite{kaifosh2025generic} acquisition and real-world usage scenarios. It includes sEMG signals from 43 subjects performing 17 gestures across 3 sessions, each comprising 7 trials recorded under similar conditions using 6 channels at 2048 Hz. We select the first session for evaluation and adopt a six-fold cross-subject cross-validation strategy, where subjects were sequentially divided into six groups. Before training, the raw sEMG signals were preprocessed using the official GrabMyo processing pipeline, where the data were segmented into 200 ms windows with a 150 ms step size. Since Crossformer~\cite{zhang2023crossformer} did not produce stable results on this dataset, we adopted GengNet~\cite{geng2016gesture} as a reliable backbone for downstream evaluation.

The results in Table~\ref{tab10} show that the proposed SASG-DA consistently improves recognition performance under the challenging cross-subject setting, outperforming all SOTA methods across three backbones (effect size $r > 0.5$). While several augmentation techniques offer limited or even negligible gains in this scenario, SASG-DA provides clear and stable improvements, indicating that generating faithful and diverse samples effectively facilitates generalization to unseen subjects. These results highlight the robustness and adaptability of our method and its potential in practical applications.

\begin{table}[!t]
\centering
\caption{Performance comparison in terms of ACC (\%) on dataset GrabMyo under cross-subject evaluations. The results report the mean and the standard deviation across folds.}
\label{tab10}
\renewcommand{\arraystretch}{0.98}
\resizebox{0.8\columnwidth}{!}{%
\begin{tabular}{lccc}
\toprule
 Aug. Method & TDCT & STCNet & GengNet  \\
\midrule
Baseline & 48.50$\pm$2.21 & 49.09$\pm$1.67 & 47.27$\pm$1.53  \\
Jittering \& Scaling  & 49.66$\pm$1.78 & 50.07$\pm$1.42 & 47.54$\pm$1.58 \\
Upsample & 48.93$\pm$1.33 & 49.23$\pm$1.36 & 46.51$\pm$1.72   \\
FreqMask & 48.23$\pm$1.60 & 48.79$\pm$1.33 & 46.26$\pm$1.70 \\
RIM & 48.48$\pm$1.31 & 49.18$\pm$1.73 & 46.59$\pm$1.74   \\
Dominant Shuffle & 47.68$\pm$0.51 & 47.52$\pm$2.24 & 45.99$\pm$1.46  \\
\midrule
Mixup & 51.00$\pm$2.61 & 52.05$\pm$2.53 & 48.68$\pm$2.33   \\
FreqMix & 48.35$\pm$1.56 & 48.04$\pm$1.85 & 47.10$\pm$1.29  \\
\midrule
DiffMix & 46.76$\pm$1.79 & 47.87$\pm$1.39 & 45.77$\pm$1.23  \\
NTD & 48.67$\pm$0.69 & 48.05$\pm$1.37 & 48.77$\pm$2.15  \\
PatchEMG & 50.01$\pm$2.61 & 51.41$\pm$3.06 & 48.63$\pm$1.54  \\
DESAM & 44.33$\pm$1.77 & 44.66$\pm$1.65 & 47.38$\pm$1.52 \\
Minority & \underline{52.65$\pm$2.73} & 52.01$\pm$2.54 & 48.12$\pm$1.33  \\
CADS & 52.64$\pm$3.26 & \underline{52.50$\pm$2.36} & \underline{48.87$\pm$1.52}  \\
\midrule
SASG-DA (Ours) & \textbf{53.53$\pm$2.35} & \textbf{54.31$\pm$2.92} & \textbf{50.06$\pm$2.03} \\
\bottomrule
\end{tabular}}
\vspace{-0.5cm}
\end{table}

\subsection{Sensitivity Analysis to Synthetic Data Size}
\begin{figure}[t]
    \centering
        \includegraphics[width=0.95\columnwidth]{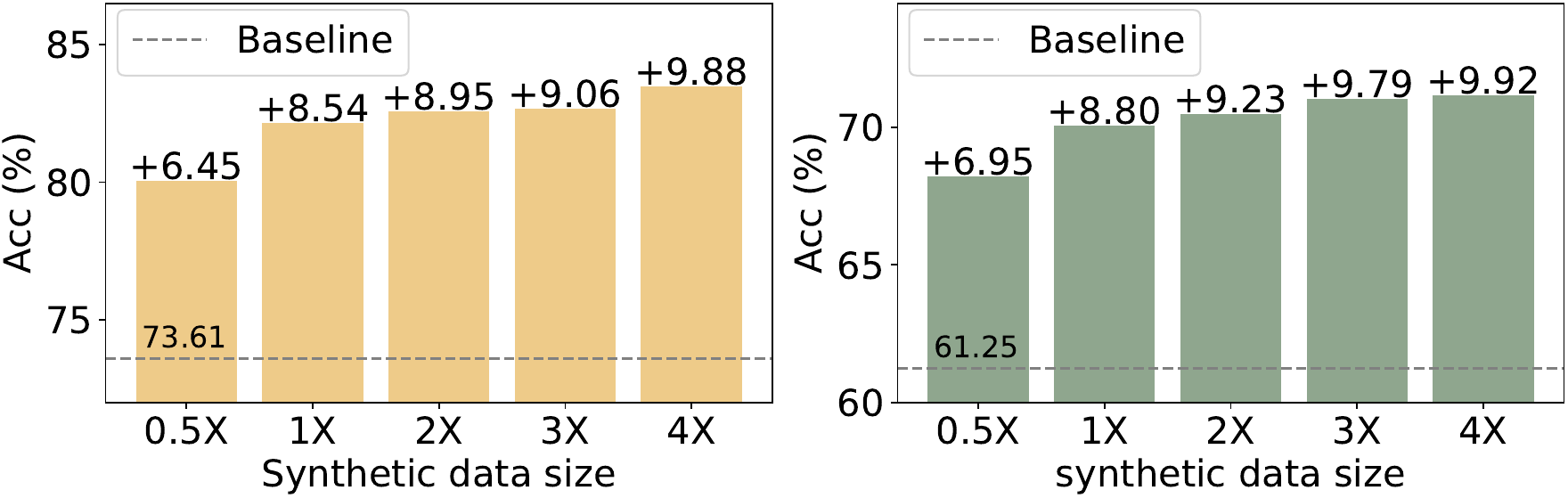}
    \vspace{-0.2cm}
    \caption{Effect of synthetic data size on classification accuracy for DB7 (left) and DB4 (right).}
    \vspace{-0.6cm}
    \label{size}
\end{figure}

We further explore the effect of the augmented size and sensitivity of the synthetic samples on model performance. To this end, we systematically increase the size of the synthetic dataset to 0.5×, 1×, 2×, 3×, and 4× the size of the original training set. As shown in Fig.~\ref{size}, experiments on both DB7 and DB4 demonstrate that the classification accuracy improves monotonically with the growth of synthetic data. Specifically, for DB7, we observe consistent gains from +6.45\% to +9.88\% compared to the baseline when enlarging the synthetic set. A similar trend is seen on DB4, where the accuracy steadily rises from +6.95\% to +9.92\% above the baseline. These results suggest that expanding the size and diversity of synthetic samples contributes to more robust model generalization. Furthermore, the consistent improvement indicates low sensitivity to the augmentation ratio, suggesting that our method preserves faithfulness to the real data distribution while providing beneficial diversity, without noticeable performance degradation within the tested range.

\subsection{Limitation and Future Work}
Although the proposed diffusion-based augmentation method effectively improves classification performance by generating faithful and diverse samples, its generation efficiency remains a limitation~\cite{yin2024one}. Due to the iterative nature of the diffusion process, producing a large amount of synthetic data can be computationally expensive and time-consuming compared to other generative models. Accordingly, SASG-DA is formulated as an offline augmentation strategy and may be less suitable for scenarios requiring on-the-fly generation of large-scale data. In future work, we plan to investigate acceleration strategies, such as distillation-based fast samplers~\cite{yin2024one}, to reduce generation costs without compromising sample diversity and faithfulness. Exploring lightweight diffusion architectures could further facilitate practical deployment in data augmentation applications.

\section{Conclusion}

Informative data scarcity and overfitting remain key challenges for real-world sEMG-based gesture recognition in HMI applications. To address them, we propose SASG-DA, a diffusion-based augmentation framework that integrates semantic representation guidance with sparse-aware semantic sampling to generate faithful and targeted diverse samples. 
By modeling and selectively sampling from a learned semantic space, SASG-DA enables flexible and controllable augmentation that enhances class faithfulness and expands coverage of underrepresented regions.
Extensive experiments on multiple public sEMG datasets show consistent improvements in recognition accuracy and generalization over existing augmentation methods. We believe SASG-DA offers a practical and principled data augmentation approach for gesture recognition and holds potential for broader biosignal and time-series classification tasks with data scarcity. Future work will explore extending this augmentation approach to cross-domain settings and other domains where semantically conditioned and diversity-aware sample generation is beneficial, as well as testing its applicability in real-time prosthetic control scenarios.

\section*{References}

\bibliographystyle{IEEEtran}
\bibliography{IEEEabrv,refs}

\end{document}